\RequirePackage{fix-cm}
\documentclass[twocolumn]{svjour3}          
\smartqed  

\usepackage{epsfig}
\usepackage{graphicx}
\usepackage{amsmath}
\usepackage{amssymb}
\usepackage{color}

\usepackage[left=2.5cm,right=2.5cm,top=3cm,bottom=3cm,a4paper]{geometry}
\usepackage[list=true]{subcaption}
\usepackage{hhline}
\usepackage{setspace}
\usepackage{subcaption}
\usepackage{rotating}
\usepackage{breakurl}

\usepackage{multirow}

\usepackage{xcolor}
\usepackage{pgfplots}
\usepackage{tikz}
\usepackage[misc,geometry]{ifsym}

\usepackage[pagebackref=true,breaklinks=true,letterpaper=true,colorlinks,bookmarks=false]{hyperref}

\newcommand{\Figref}[1]{Figure~\ref{#1}}

\newcommand{\figref}[1]{Fig.~\ref{#1}}
\newcommand{\tabref}[1]{Table~\ref{#1}}
\newcommand{\eqnref}[1]{Eq.~(\ref{#1})}
\newcommand{\secref}[1]{Sec.~\ref{#1}}

\newcommand{\comment}[1]{{}}

\newcommand{\ie}{\textit{i.e.}}

\newcommand{\etal}{\textit{et al.}}

\newcommand{\GM}[1]{\textcolor{blue}{{[GM: #1]}}}
\newcommand{\JS}[1]{\textcolor{magenta}{{[JS: #1]}}}

\journalname{IJCV preprint}
\begin{document}

\title{Refining Geometry from Depth Sensors using\\ IR Shading Images}

\author{Gyeongmin Choe \and Jaesik Park	\and Yu-Wing Tai \and In So Kweon}

\institute{G. Choe \and I.S. Kweon \at
School of Electrical Engineering, 
Korea Advanced Institute of Science and Technology (KAIST), \\
Daejeon, Republic of Korea.\\
\email{gmchoe@rcv.kaist.ac.kr,~iskweon77@kaist.ac.kr}
\and
J. Park \at
Intel Labs, Santa Clara, CA \\
\email{jaesik.park@intel.com}
\and
Y-W. Tai \at 
SenseTime Group Limited, Hong Kong China \\
\email{yuwing@sensetime.com}
}

\date{Received: date / Accepted: date}

\maketitle

\begin{abstract}
We propose a method to refine geometry of 3D meshes
from a consumer level depth camera, e.g. Kinect, by exploiting shading cues captured from an infrared
(IR) camera. A major benefit to using an IR camera
instead of an RGB camera is that the IR images captured are narrow band images that filter out most undesired ambient light,
which makes our system robust against natural indoor illumination. Moreover,
for many natural objects with colorful textures in the visible spectrum, the subjects
appear to have a uniform albedo in the IR spectrum. Based on our
analyses on the IR projector light of the Kinect, we define a
near light source IR shading model that describes the captured intensity
as a function of surface normals, albedo, lighting direction, and distance
between light source and surface points. To resolve the ambiguity in our model
between the normals and distances, we utilize an initial 3D mesh from the Kinect
fusion and multi-view information to reliably estimate surface details that
were not captured and reconstructed by the Kinect fusion. Our approach directly operates
on the mesh model for geometry refinement. We ran experiments on our algorithm for geometries captured by both
the Kinect I and Kinect II, as the depth acquisition in Kinect I is based on
a structured-light technique and that of the Kinect II is based on a time-of-flight (ToF) technology.
The effectiveness of our approach is demonstrated through several challenging
real-world examples. We have also performed a user study to evaluate the quality of the mesh models before and after our refinements.
\end{abstract}

\section{Introduction}
\label{intro}
Over the past few years, Microsoft Kinect\footnote{\url{http://www.microsoft.com/en-us/kinectforwindows/}}
 has become a popular input
device in depth map acquisition for human pose recognition~\cite{Shotton11cvpr}, 3D
reconstruction~\cite{KinectFusion}, robotics~\cite{kerl13iros} and many other 
applications~\cite{KinectIdentity}. The Kinect I utilizes active range sensing by
projecting a structured light pattern, \ie~a speckle pattern, on a scene in the infrared
(IR) spectrum\footnote{Strictly speaking, it captures a range between 800$nm$ and 2500$nm$, which belong to the near infrared band. For simplicity, we abbreviate the band as the IR band in this paper.}. By analyzing the displacement of the speckle pattern, a depth map of
the scene can be estimated. In the Kinect II, although the underlying technique for depth map acquisition is based on a time-of-flight (ToF)
technology, the Kinect II still retains the IR projector and IR camera for the Kinect to capture 
images under dark environments.

\bgroup
\def\arraystretch{1.05}
\begin{table}
\centering
\resizebox{1.0\linewidth}{!}{
\hspace{-2mm}
\begin{tabular}{
|@{\hspace{1mm}}c@{\hspace{1mm}}|
|@{\hspace{1mm}}c@{\hspace{1mm}}
|@{\hspace{1mm}}c@{\hspace{1mm}}
|@{\hspace{1mm}}c@{\hspace{1mm}}
|@{\hspace{1mm}}c@{\hspace{1mm}}
|@{\hspace{1mm}}c@{\hspace{1mm}}
|@{\hspace{1mm}}c@{\hspace{1mm}}|}\hline
\begin{turn}{90} \end{turn} & 
\begin{turn}{90} publication\end{turn} & 
\begin{turn}{90} color camera\end{turn} &
\begin{turn}{90} depth camera\end{turn} & 
\begin{turn}{90} auxiliary lights\end{turn} &
\begin{turn}{90} light model\end{turn} & 
\begin{turn}{90} variables \end{turn}
\begin{turn}{90} to be optimized~~~~\end{turn}\\ \hline \hline
Nehab~\cite{ravi_sig05}	& 05 &$\checkmark$ & $\checkmark$ & $\checkmark$ & DP & vertex position \\
Hernandez~\cite{Hernandez08pami} & 08 & $\checkmark$ & & $\checkmark$ & DP & vertex position \\
Wu~\cite{yasuMultiview} & 11 & $\checkmark$ & & & SH & vertex position \\
Zhang~\cite{Zhang12cvpr} & 12 & $\checkmark$ & $\checkmark$ & $\checkmark$ & DP	& depth map \\
Park~\cite{Park13ICCV} & 13 &  $\checkmark$ & & $\checkmark$ & DP & displacement map \\  
Han~\cite{Han13ICCV} & 13 & $\checkmark$ &  $\checkmark$ &  & QF & surface normal \\
Yu~\cite{Yu13cvpr} & 13 & $\checkmark$ & $\checkmark$ & & SH & surface normal \\
Wu~\cite{Wu14tog} & 14 & $\checkmark$ & $\checkmark$ & & SH & vertex position \\ 
Zollhofer~\cite{zollhofershading} & 15 & $\checkmark$ & $\checkmark$ & & SH & position of voxel \\
Or-El~\cite{or2015rgbd} & 15 & $\checkmark$ & $\checkmark$ & & SH & depth map \\ \hline \hline
Bohme~\cite{Bohme10cviu} & 10 & & $\checkmark$ & & NP & depth map  \\ 
Haque~\cite{Haque14cvpr} & 14 & & $\checkmark$ &$\checkmark$ & DP &  depth map \\ 
Chatterjee~\cite{chatterjee2015photometric} & 15 & & $\checkmark$ & $\checkmark$ & DP & surface normal \\ 
\bf{Ours} & -- & & $\checkmark$ & $\Delta$ & NP & vertex displacement \\ \hline
\end{tabular}
}
\vspace{0mm}
\caption{Representative approaches for geometry refinement via shape from shading or photometric stereo. Simplified notations for light model indicate; DP: Distant point light, SH: Spherical harmonics, QF: Quadratic function, NP: Near point light. Our method is easily applicable to commercial depth sensors using the near IR band~(\tabref{table:Sensors}). In addition, the optimized variable of our method is a 1D displacement for each vertex, which makes our optimization variable simpler than that of other methods.}
\label{table:summary}
\end{table}
\egroup

The success of the Kinect relies heavily on the usage of the narrow-band IR camera, which filters out most of the undesired ambient light, making the depth acquisition robust to natural indoor illumination. Although the 
IR camera is one of the key components to the success of the Kinect, after the depth acquisition,
these IR images are discarded and not used in any post-processing applications. 
In this paper, we show that the IR camera of the Kinect is not only useful in
the depth measurement, but also useful for capturing 
shading cues of a scene that allow higher quality reconstruction than the Kinect
fusion~\cite{KinectFusion}, which only uses the estimated depth map for 3D reconstruction.

We analyzed the properties of the light emitted by the IR projector of the Kinect and found that
the projector light can be approximately modeled by a near point light source with the 
light falloff property \cite{liao2007light}, where its illumination falls off with distance according to the inverse square law. 
With the Lambertian BRDF assumption about the scene materials in the captured IR spectrum, 
we define a near point light IR shading model that describes the captured intensity as a function of 
surface normals, albedo, lighting direction, and distance between a light source and surface points. 
The proposed model has an ambiguity between the normals and distance estimations using a single shading
image. Therefore, we utilize an initial 3D mesh from the Kinect fusion and shading images from 
different view points. Our approach operates directly on the 3D mesh and optimizes the geometry 
refinement process subject to the shading constraint of our model. The result is a high quality 
mesh model that captures surface details, which were not reconstructed by the Kinect fusion. 
Thanks to the usage of the Kinect IR camera, our approach is also robust to indoor 
illumination and works well in both dark rooms and natural lighting environments. Furthermore, we have also
found that for many materials with colorful albedo in the visible spectrum, the objects appear to have an 
uniform albedo in the IR spectrum. This observation allows us to use a simple technique
to estimate surface albedo with reliable accuracy. Our approach does not require any additional cameras 
nor complicated light setups, making it useful in practical scenarios as an add-on to enhance reconstruction 
results from the Kinect fusion. Since the speckle pattern in the Kinect I is hardwired, we use a 
broad spectrum light bulb to approximate the IR projector light of the Kinect I with calibration. In the
Kinect II, which uses a ToF technology, the inherent IR light source allows us to get a shading image without the additional light bulb.

This paper extends our previous work published in~\cite{Choe14cvpr}. Specifically, the major benefits of using IR shading images for geometry refinement are further analyzed. We have also provided additional technical details in the albedo estimation and geometry optimization. To demonstrate the flexibility of our algorithm, results using only a single depth map and an IR image pair is also included. Similar to the Kinect I, the sensor characteristics of the Kinect II are also covered and the refined results from both sensors are displayed. To verify the effectiveness of our method, we conduct both a quantitative error measure and a qualitative user study. The rendered shading images from the meshes of the Kinect fusion and our method are compared to the input IR shading image. It measures how accurately  our refined mesh models follow the photometric cues of the IR shading image. The user study also demonstrate improvements in terms of the visual quality of our refined mesh model.

\section{Related Works}
In the recent decade, depth measurement devices, such as the Kinect or ToF cameras, have allowed users to easily acquire a depth map of the scene at a low cost. However, the depth map usually contains holes and noisy measurements, which makes it less useful when
a high quality depth map is required. Utilizing the additional RGB image, methods in~\cite{Yang07cvpr,Dolson10cvpr,jaesik11,jaesik14,Shen13cvpr} define a smoothness cost according to the image structures in the RGB image for depth map refinement, but their approaches do not use any shading information to potentially improve the depth quality. 

Many literatures utilize shading or surface normal cues for the enhancement of rough geometry. Nehab~\etal~\cite{ravi_sig05} refines a depth map by enforcing orthogonality between the surface gradient of the depth and surface normal acquired from photometric stereo~\cite{horn1978,liao2007light,higo2009hand}. Recently, Haque~\etal~\cite{Haque14cvpr}, extends the work of~\cite{ravi_sig05} by utilizing IR images instead of color images.
Work in~\cite{Lu10cvpr} utilizes a giga-pixel camera to estimate ultra high resolution surface normals from
photometric stereo to refine a low resolution depth map captured by using a structured light. B¨ohme~\etal~\cite{Bohme10cviu} uses
shading information to improve depth map from a ToF camera. In~\cite{Zhang12cvpr,Okatani12cvpr}, they use normals from photometric
stereo to refine a depth map with additional consideration to depth discontinuities~\cite{Zhang12cvpr} and the first-order derivative of surface
normals~\cite{Okatani12cvpr}. Recent works by~\cite{Han13ICCV,Yu13cvpr,Wu14tog} propose to use shape-from-shading~\cite{HornBook,Ikeuchi81ai} from an RGB image to estimate surface details for depth map refinement. In~\cite{Suwajanakorn14eccv}, high quality facial shape is generated using photometric cues of the color video sequences. In~\cite{Shi143DV}, photometric normals are obtained from a collection of internet photos with a linear approximation of the camera response functions, and then 3D shape of the object is refined.

\begin{figure}[t]
    \centering
    \includegraphics[width=0.995\linewidth]{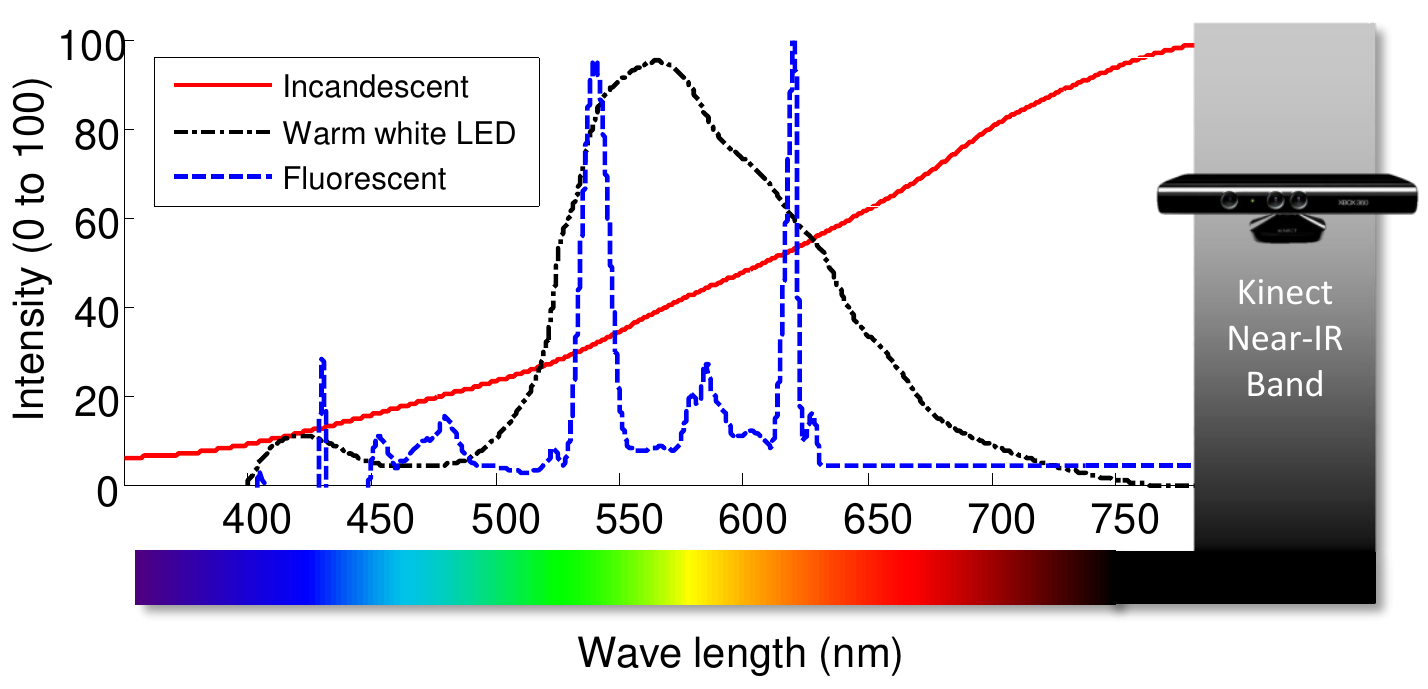}
    \caption{Frequency responses of several light sources~\cite{Bellia11environment}. Since general indoor lightings such as fluorescent bulbs or LEDs emit only visible light and incandescent light emits large amount of near-IR light, the IR camera of the Kinect only senses incandescent light under complex indoor lighting conditions. This makes our algorithm work robustly with our simple lighting model.}
\label{fig:FreqLight} \vspace{0in}
\end{figure}

\begin{figure*}
    \centering
    \begin{tabular}{@{}c@{ }c@{ }c@{ }c}
        \includegraphics[width=1\linewidth]{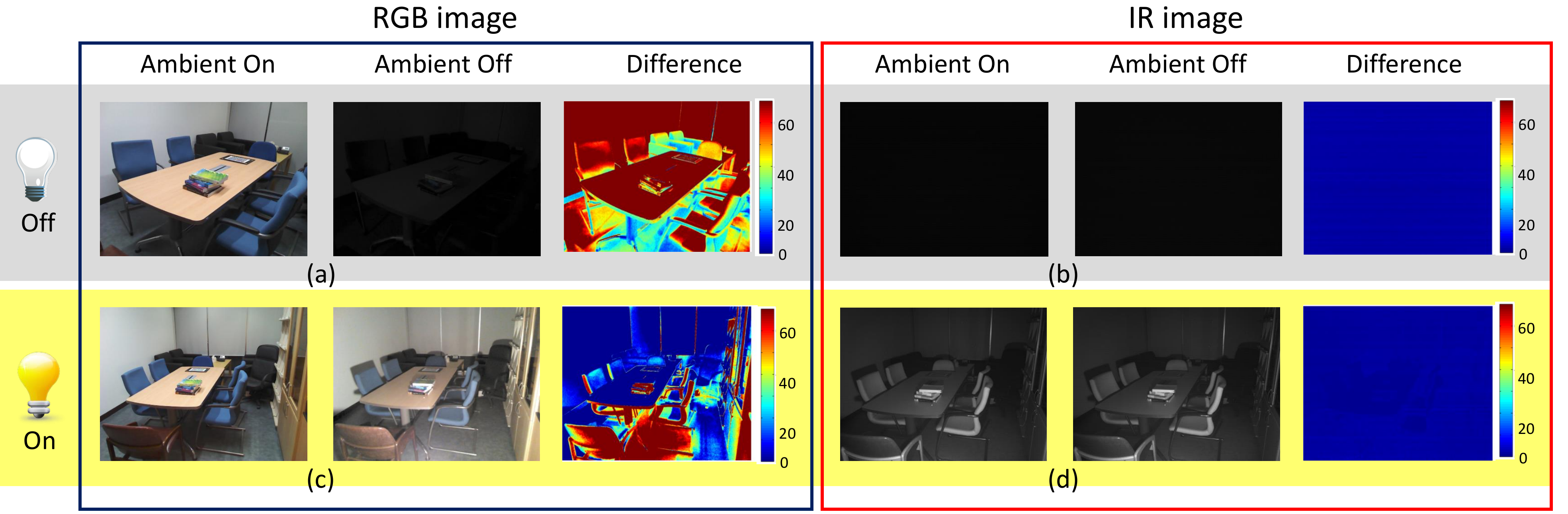}
    \end{tabular}
    \caption{Invariability of Kinect IR images under different lighting conditions. (a) RGB images under ambient light and dark room. (b) The corresponding Kinect IR images of (a). (c) RGB images under ambient light and dark room with an additional wide spectrum light source. (d) The corresponding Kinect IR images of (c). The difference images are shown on rightmost columns of each image pairs. Enormous differences are observed in the RGB image pairs while the IR image pairs are almost identical.}
\label{fig:ColorIRSubtract} 
\end{figure*}

In 3D mesh refinement methods, the start typically consists of a rough 3D mesh model estimated by using stereo matching~\cite{Seitz00cvpr_mvstereo}, visual hull~\cite{Hernandez08pami}, structure from motion~\cite{LonguetHiggins81nature}, or Kinect fusion~\cite{KinectFusion}. Similar to 2D depth map refinement, Hernandez~\etal~\cite{Hernandez08pami} demonstrate a two-way stage that estimates light directions and refines mesh model to have an estimated surface normal direction. Lensch~\etal~\cite{Lensch03tog} introduce a generalized method for modeling non-Lambertian surfaces by using wavelet-based BRDFs and use it for mesh refinement. Vlasic~\etal~\cite{Vlasic09tog} integrate per-view normal maps into partial meshes, then deforms them using thin-plate offsets to improve the alignment while preserving geometric details. Wu~\etal~\cite{yasuMultiview} use the multi-view stereo to solve the shape-from-shading ambiguity. They demonstrate high-quality 3D geometry under arbitrary illumination but assume the captured objects contain only a single albedo. Park~\etal~\cite{Park13ICCV} refine 3D mesh in parameterized space and demonstrate state-of-the-art quality in geometry refinement results using normals from photometric stereo.
Recently, Delaunoy~\etal~\cite{Delaunoy14cvpr} propose a dense 3D reconstruction technique that jointly refines the shape and the camera parameters of a scene by minimizing the photometric reprojection error between a generated model and the observed images. Also Fanello~\etal~\cite{Fanello14tog} propose a method for recovering the dense 3D structures of human hands and faces. They use hybrid classification-regression forests to learn how to map near infrared intensity images to absolute, metric depth in real-time.


Comparing our work to the previous works, especially for the 3D mesh refinement methods, most of them utilize photometric stereo to estimate normal details. Although high-quality surface details can be estimated by photometric stereo, as demonstrated in the experimental setting
in~\cite{Hernandez08pami,Vlasic09tog,Park13ICCV}, they require control over the environment's illumination. In contrast, our work utilizes the Kinect IR camera, which makes our approach robust to natural indoor illumination as shown in \figref{fig:FreqLight}. In addition, we define a near point light shading model that fits perfectly to our problem setting to utilize instead of a directional light source for normal estimation. Since our work directly operates on the mesh model, our approach is also efficient and effective in mesh model refinements.

\begin{figure}[t]
    \centering
    \includegraphics[width=0.995\linewidth]{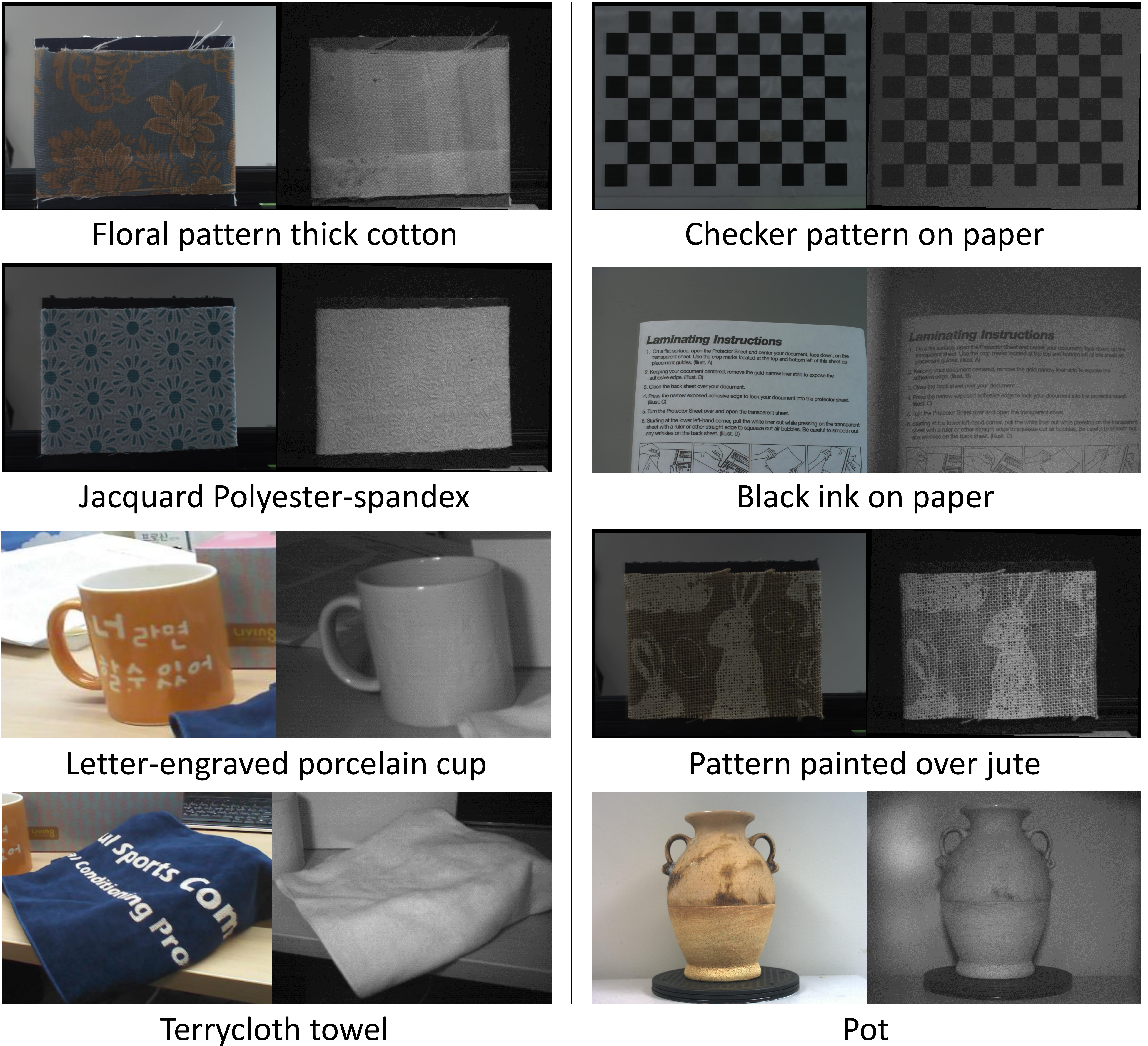}
    \caption{Image pairs of different materials in visible and IR spectrum. Left: Color pigments in visible spectrum are invisible in IR spectrum. Right: Black inks are visible in both visible and IR spectrum.}
\label{fig:AlbedoVariance} 
\end{figure}


\section{IR Shading Images}
In this section, we first analyze the IR images captured by the Kinect I and Kinect II. The inverse square law property of the IR light source is verified. 
The benefit of using IR images for simplifying albedo estimation in a scene is also analyzed. After that, we
define our near point source IR light shading model. A radiometric calibration technique based on our IR shading model is also presented.

\vspace{-2mm}
\subsection{Kinect IR Images}
\label{sec:KinIRImg}
We verify the invariability of Kinect IR images under different indoor lighting conditions.
In \figref{fig:ColorIRSubtract}, we block the Kinect IR projector and then capture IR
images under ambient light and dark room environment. The RGB image pairs in (a) show enormous intensity
differences under the two different lighting conditions, but the IR image pairs in (b) are almost identical.
Next, we put a wide spectrum light source and then capture the RGB and IR images again under the same
ambient light and dark room environment. Again, enormous intensity differences are shown in RGB image pairs in (c), while the IR image pairs in (d) have almost no difference. This example shows that common indoor lighting conditions do not cover the IR spectrum captured by the Kinect IR camera. Unless a wide spectrum light source
is presented in a scene, the Kinect IR images is unaffected by ambient lighting.

In addition to the invariant indoor ambient light characteristic, the chromatic variations of textures in the visible spectrum appear to have a uniform albedo in the IR spectrum. In~\figref{fig:AlbedoVariance}, we capture the same scene with a color camera and an IR camera. 
Textures on the mug, the towel, and the fabric appear to have uniform color in the IR spectrum, whereas, black ink is visible in both the visible and IR spectrum. This property is further analyzed by Salamati \etal~\cite{Salamati09cic}. They captured images of many different types of materials in the visible and near IR spectrum. The paper analyzed luma, intensity, and color information of images in the two different spectrum and revealed that many pigments used to colorize materials appear to be transparent in the near IR spectrum. Based on this analysis, we can simplify the albedo estimation by assuming that the same materials have the same albedo in the IR spectrum. This allows us to impose a smoothness regularization in the albedo estimation.

\begin{figure}[t]
\centering
\begin{tabular}{@{}c@{ }c@{}c@{}c@{}}
\includegraphics[width=0.14\linewidth]{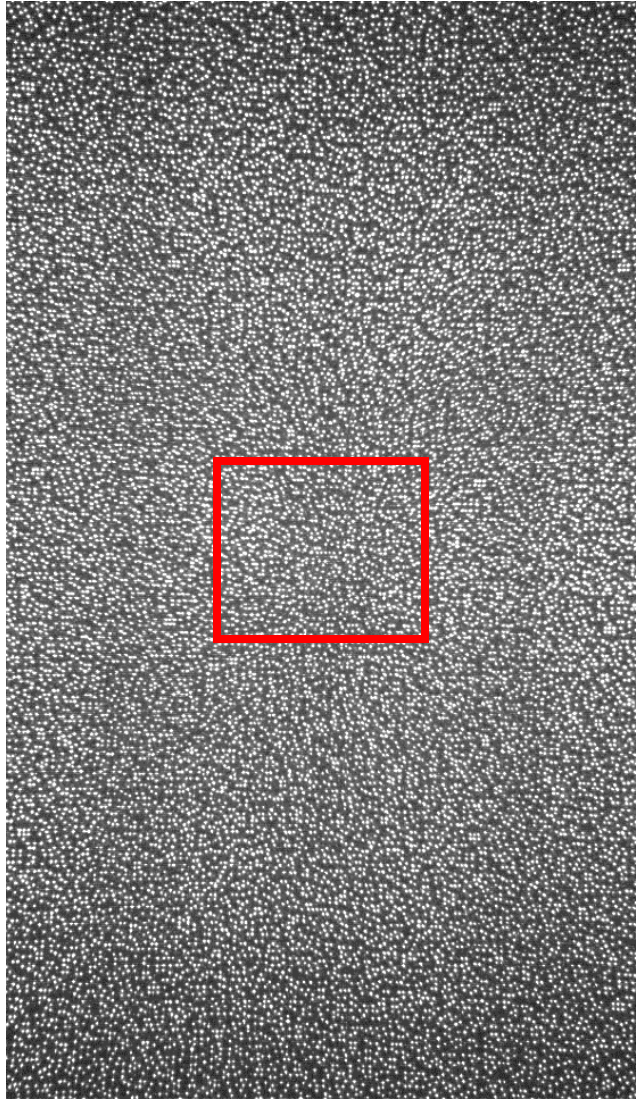}
&\includegraphics[width=0.14\linewidth]{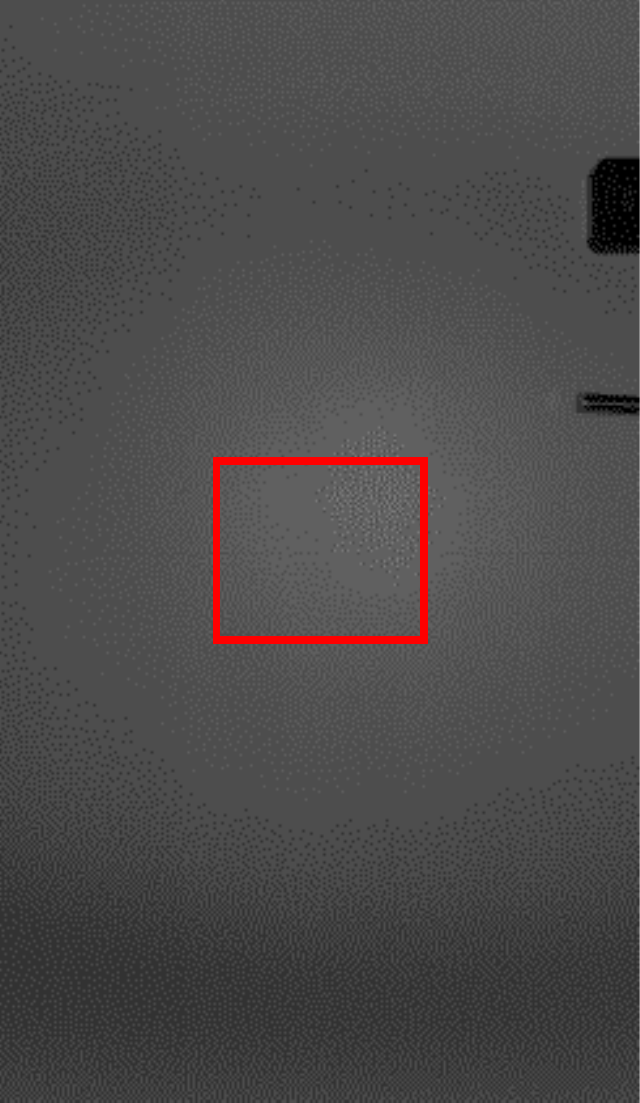}
&\includegraphics[width=0.36\linewidth]{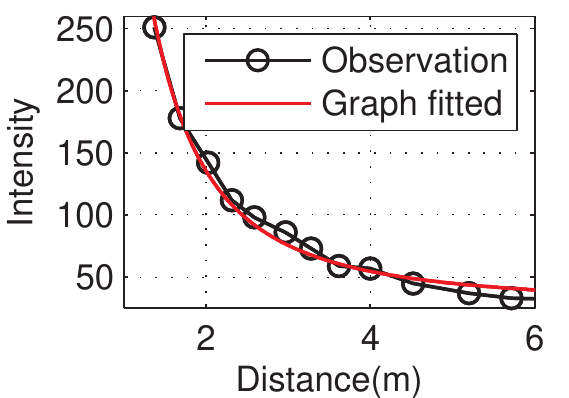}
&\includegraphics[width=0.36\linewidth]{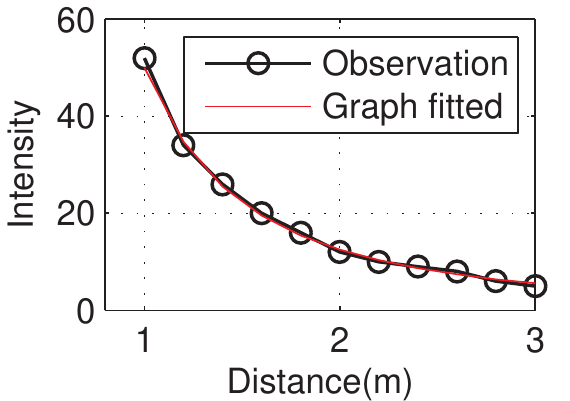}\\
\small{(a)}&\small{(b)}&\small{(c)}&\small{(d)}\\
\end{tabular}
\caption{Validation of the inverse square law. (a, b)  Region of Interest (ROI) in IR image of Kinect I and II respectively. (c, d) Various images at different depths are captured and the median intensity within each ROI is plotted for Kinect I and II, respectively. The observed intensity falls off with increasing distance and the falloff rate follows the inverse square law.}
\label{fig:inverse_squarelaw}
\end{figure}

Our third analysis verifies the inverse square law property of the Kinect IR projector, a near point light source. We capture
IR images at different distances of a white wall. Since we capture various images at different depths, we have observed that the number of total pixels grows too much for curve fitting (Each ROI contains 40k pixels (200 x 200), at least 10 images are used, results in 400k pixels). Therefore, for an efficient computation, we obtained the median intensity that is the representative intensity value for each image. \figref{fig:inverse_squarelaw} shows the captured
IR image\footnote{The IR image is radiometrically calibrated.} and the region of median intensity
with the red box. The decay of observed intensity follows the inverse square law.

\subsection{Near IR Light Shading Model}
\label{sec:LightModel}
Following the analyses from the previous section, we define the observed pixel intensity $I$ in the IR image as follows:
\begin{equation}
I_i = \bigg(\frac{c\rho_i}{d_i^{2}}(\mathbf{n}_i\cdot \mathbf{l}_i ) + I_{Ambient}\bigg)^\gamma,
\label{eq:IR_projection_model_2}
\end{equation}

where $i$ is the index of a 2D pixel (which will also be used as an index for the corresponding 3D vertex in \secref{sec:geometryrefinement} ), $c$ is the global brightness, $\rho$ is the albedo of the surface,
 $\mathbf{n}\in\mathbb{R}^3$ is the surface normal,
$\mathbf{l}\in\mathbb{R}^3$ is the lighting direction, and $d$ is the distance between the surface
point and the light source. $\gamma$ is the coefficient of the nonlinear radiometric parameter. Here, we assume the captured materials in the IR spectrum follow the Lambertian BRDF model. The inverse 
square term $d$ is added to account for the light falloff property along with the distance.

Since the effect of indoor ambient lights to the IR
image is subtle, we regard $I_{Ambient}=0$. Since different pairs of $d$ and $\boldsymbol{n}$ can produce identical intensity assuming
known albedo and lighting direction, we utilize the initial mesh from the Kinect and multiple
view point information to resolve this ambiguity. In \secref{sec:geometryrefinement}, we will show that this shading model is an effective constraint
for geometry refinement.


\begin{figure}[t]
\centering
\begin{tabular}{@{}c@{ }c@{}c@{}c@{}}
\includegraphics[width=0.15\linewidth]{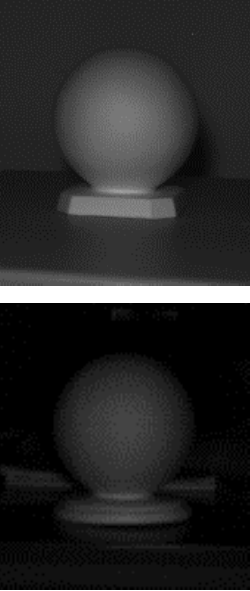}
&\includegraphics[width=0.38\linewidth]{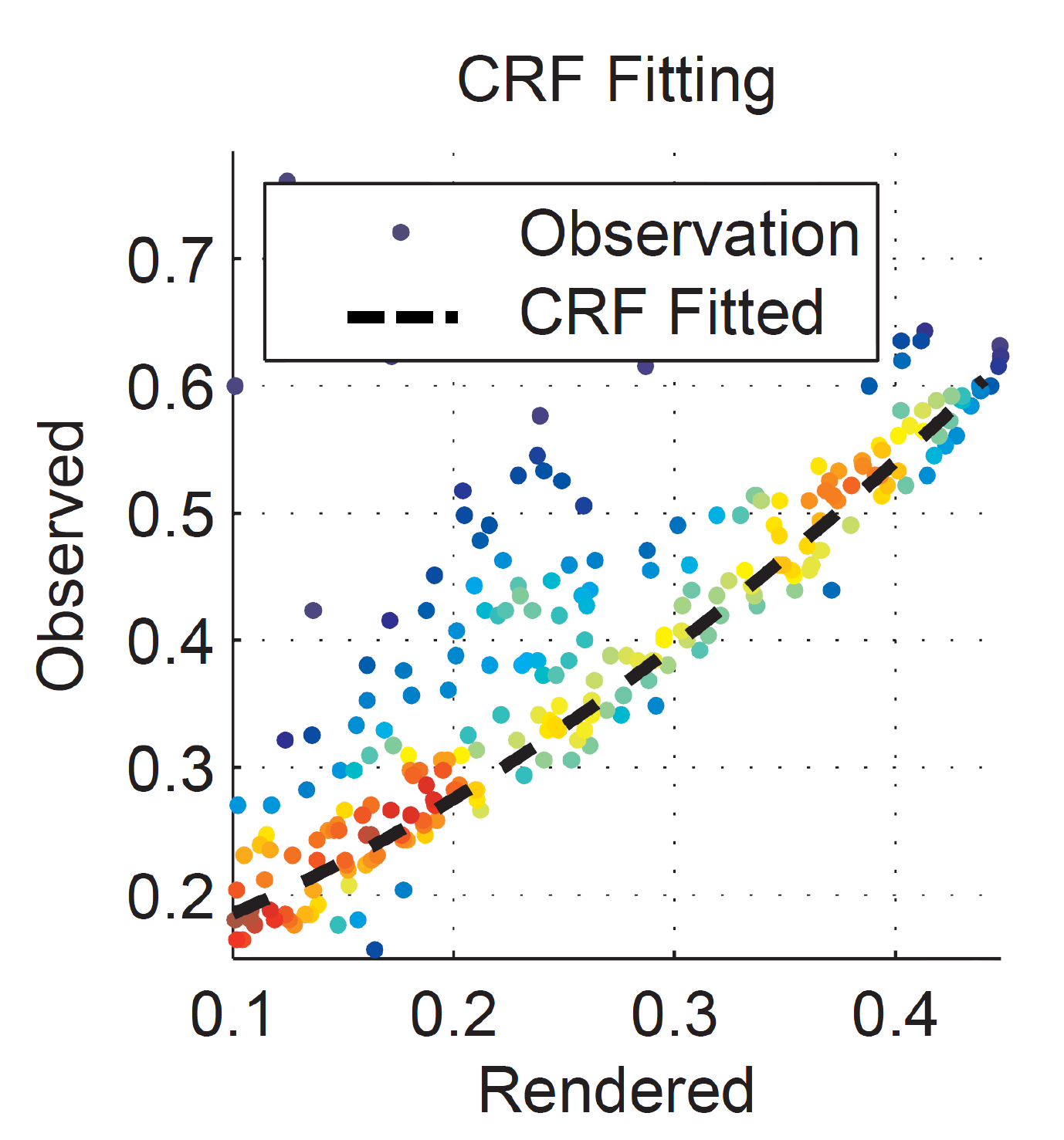}
&\includegraphics[width=0.46\linewidth]{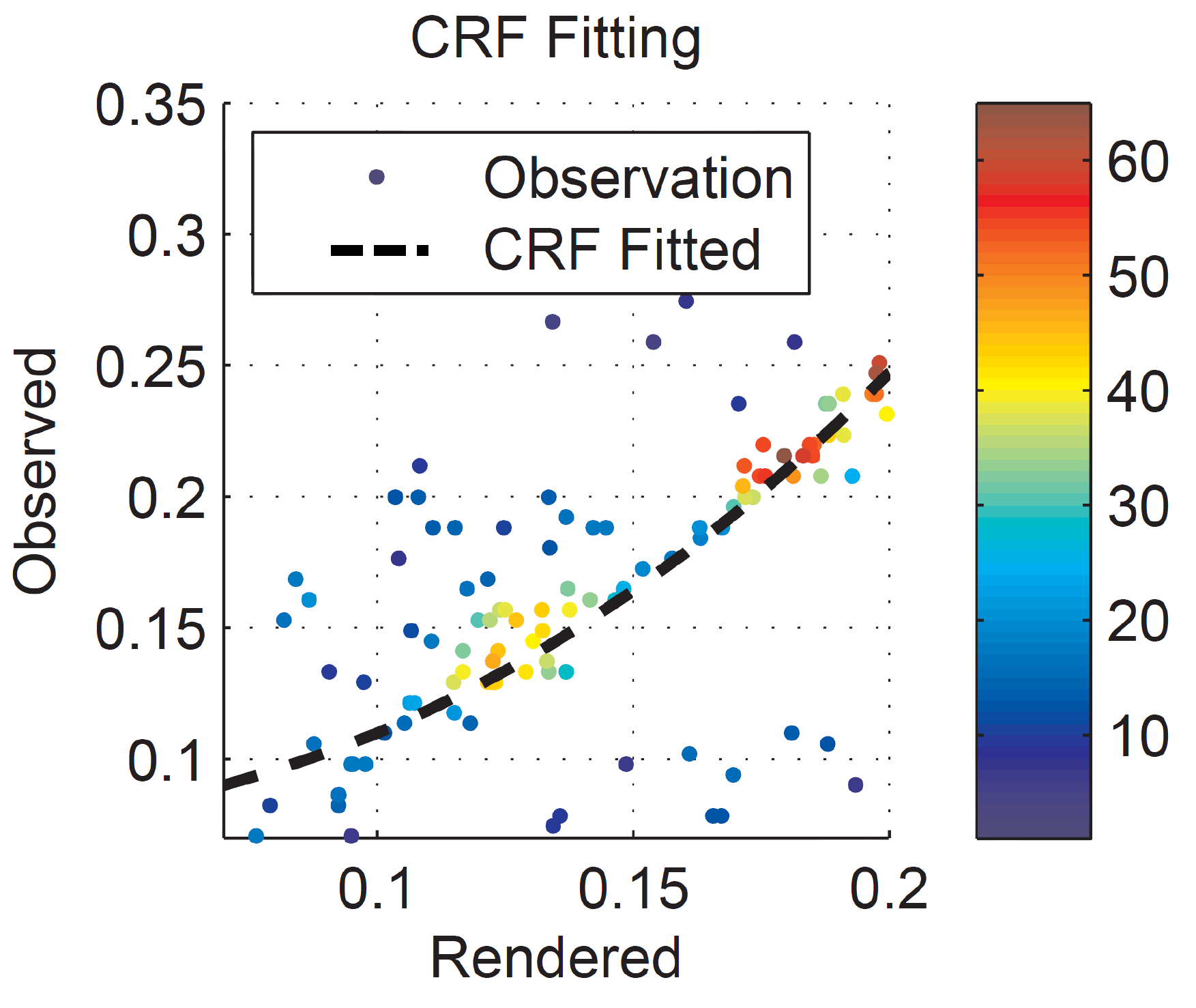}\\
\small{(a)}&\small{(b)}&\small{(c)}\\
\end{tabular}
\caption{Estimation of camera response function (CRF) of Kinect IR camera. (a)  IR image of Kinect I and Kinect II of the spherical object for calibration. (b, c) Curve fitting of CRF estimation for both sensors, respectively. The x-axis shows the rendered intensities from the base mesh and the y-axis shows the measured intensities from (a). The color-coded points show the density of the points. Note that the ratio of pixels with less than 0.05 pixel error is 76 and 78\%, respectively.}
\vspace{0in}
\label{fig:calibration}
\end{figure}

\begin{figure}
    \centering
    \begin{tabular}{@{}c@{ }c@{ }c@{ }c}
        \includegraphics[width=0.90\linewidth]{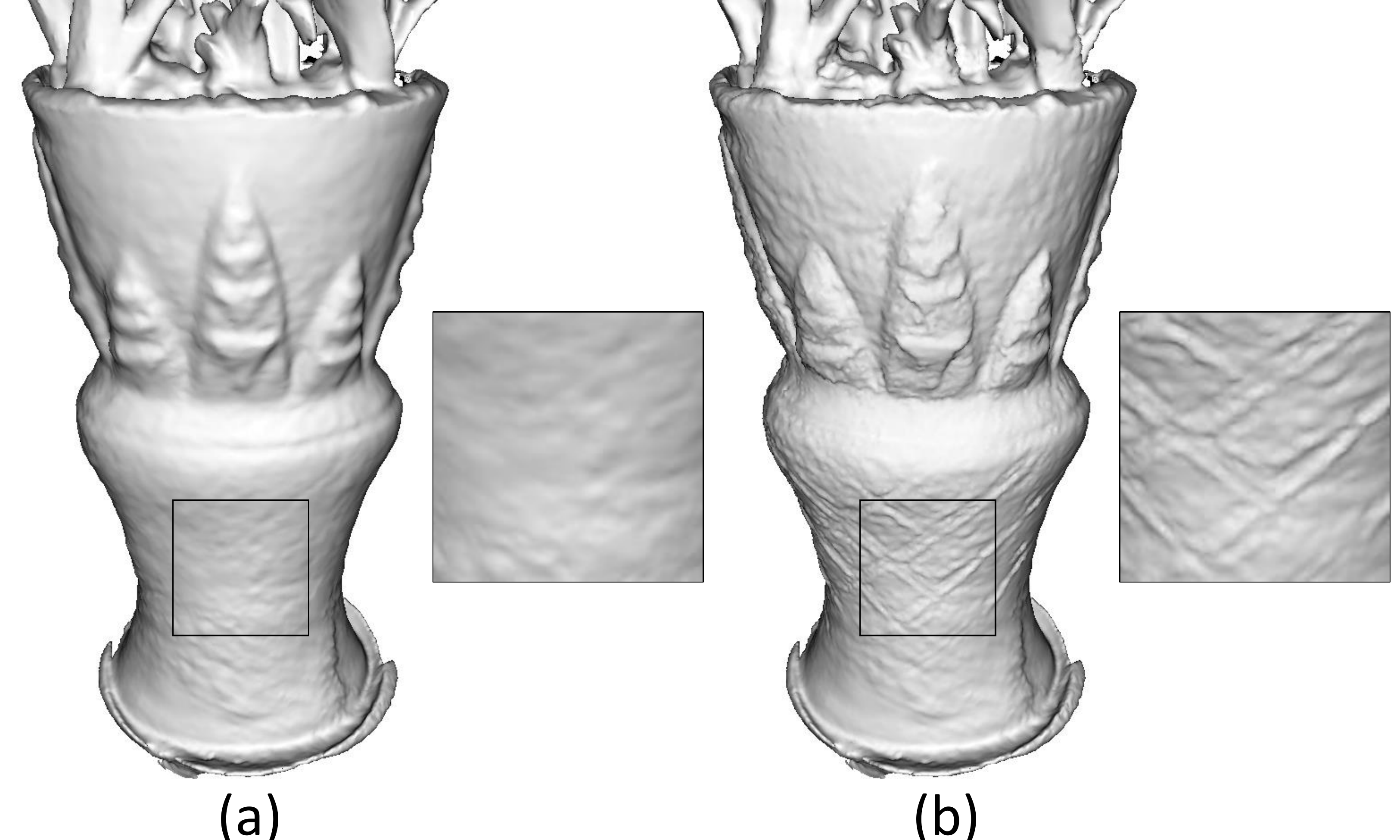}
    \end{tabular}
\caption{The validation of radiometric calibration step. (a) Our refinement result using the original IR shading image. (b) Our refinement result using radiometrically calibrated IR shading image. 
}
\label{fig:CompareGamma}\vspace{-0.1in}
\end{figure}
\subsection{Radiometric Calibration of IR camera}
\label{sec:Radiometric}
We note that the responses of the Kinect IR camera is not strictly linear to the luminance of incoming light.
Therefore, we need to radiometrically calibrate the Kinect IR camera. In previous works for radiometric
calibration~\cite{Grossberg04pami}, multiple different exposure images can be easily captured for calibration. However, the
Kinect IR camera can only capture a single exposure image. In addition, there is no calibration pattern
for IR camera calibration. Here, we propose a radiometric calibration method which makes use of
multiple photometric observations of a known geometry to estimate the camera response function (CRF)
of the Kinect IR camera.

We use a white Lambertian sphere as shown in \figref{fig:calibration} (a) for our calibration. The white sphere has a known geometry
and complete observation of surface normals in every direction. We use the Kinect fusion to obtain a base mesh of the sphere, and
then capture the IR shading images of the sphere. Since the geometry, the distance, the lighting direction, and the albedo are known
for this calibration object, we can synthetically render a predicted observation using \eqnref{eq:IR_projection_model_2}. By comparing
the measured intensities, $I_{obs}$, with the predicted intensities, $I_{ren}$, we can estimate the CRF, $f$, by fitting a curve
that minimizes the least square errors, $||I_{obs}-f(I_{ren})||^2$, as illustrated in \figref{fig:calibration} (b) and (c). Here, we assume that
$f$ is a gamma function where $I_{obs}=(I_{ren})^\gamma$. The RANSAC algorihm~\cite{fischler1981random} with 1000 sample points  and  iterations is used for robust fitting. In our estimation, we find that the gamma value is approximately equal to 0.8 for the Kinect I and 0.87 for the Kinect II.

To validate the effectiveness of the radiometric calibration step, we provide an additional experiment. First, we prepare two sets of input images that are processed with or without gamma correction. Second, we individually perform mesh refinement using different image sets. Here, the same parameters are used for the comparison. As shown in \figref{fig:CompareGamma}, the refined mesh looks nicer when our radiometric calibration step is applied a priori.

\section{Geometry Refinement}
\label{sec:geometryrefinement}

This section includes our vertex optimization method for geometry refinement. We begin this section with mesh preprocessing and surface albedo estimation of the geometry. After that, we describe our mesh refinement process.

We denote $\mathbf{x}_{i}\in\mathbb{R}^3$, the $i$-th vertex on the base mesh, $\mathbf{x}_{j}\in N(\mathbf{x}_{i})$, the neighboring vertices that
directly connect to $\mathbf{x}_{i}$, $\mathbf{K}\in\mathbb{R}^{3\times3}$ is the intrinsic camera matrix for the IR cameras in the depth sensors and $\mathbf{P}_m\in\mathbb{R}^{3\times4}$ are the extrinsic projection matrices of the camera poses from the $m$-th view.
The image coordinate $\mathbf{u}_{i,m}\in\mathbb{R}^2$ of vertex $\mathbf{x}_i$ that is projected on the $m$-th view is computed, $\mathbf{u}_{i,m}=\mathbf{K}\mathbf{P}_m\mathbf{x}_i$. We also define
$V_{i,m}$ which represents the visibility of $\mathbf{x}_i$ on the $m$-th view. \Figref{fig:vertice_projection} shows an example of vertices projection on
one of the input shading images.

\begin{figure}
    \centering
    \begin{tabular}{@{}c@{ }c@{ }c@{ }c}
        \includegraphics[width=0.995\linewidth]{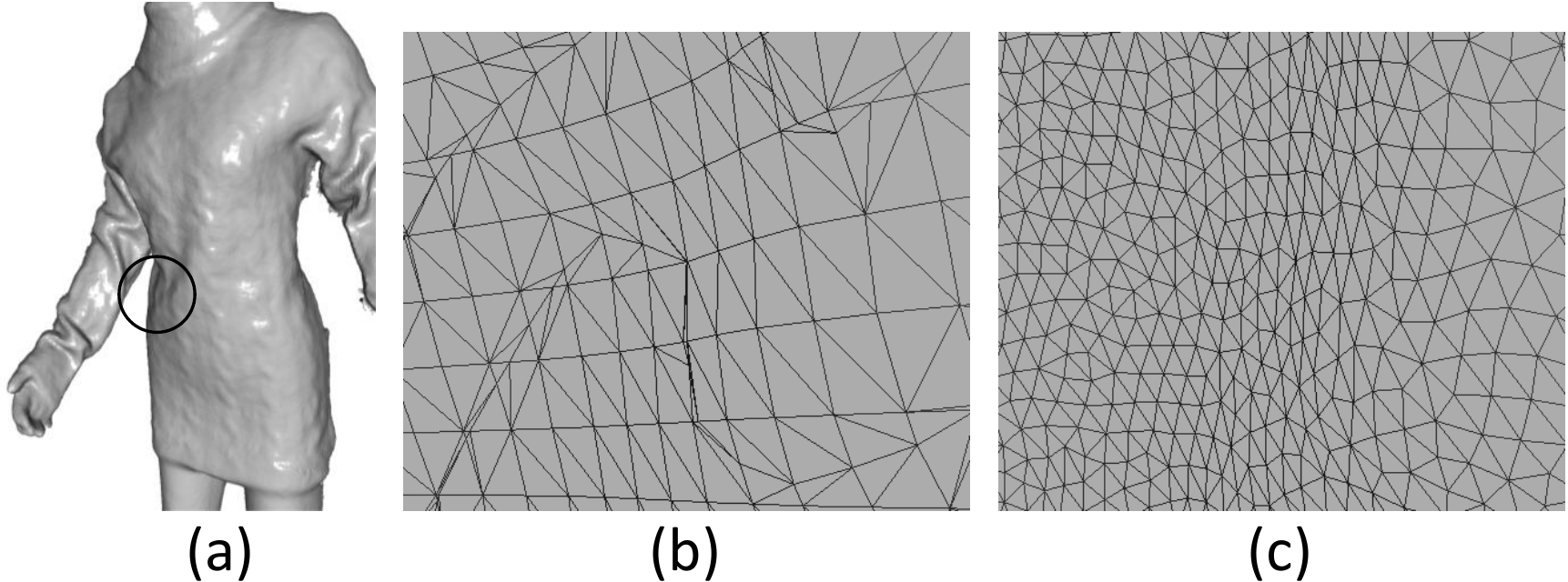}
    \end{tabular}
\caption{Mesh comparison of before and after remeshing. (a) Region of interest (ROI) of mesh (b) Initial mesh from Kinect fusion. (c) Our mesh after remeshing. Since the mesh is more clear and dense than (b), we can optimize the displacement of vertices to recover fine details effectively.}
\label{fig:Remeshing}
\end{figure}

\subsection{Mesh Preprocessing}
\label{sec:preprocessing}
Our mesh optimization controls vertex positions along with surface normal directions. For better convergence of the optimization and avoidance of mesh flipping, the initial mesh needs to be smooth enough and the vertices be uniformly distributed. 

If a rough mesh is obtained from Kinect fusion~\cite{KinectFusion}, the mesh is already smooth because the integrated depth in a voxel grid suppresses depth noise. In this case, We only apply the remeshing technique~\cite{Surazhsky03euro} to resample vertex positions uniformly as shown in \figref{fig:Remeshing}. The number of vertices are set to be about 100-200K which does not affect the initial geometry while allowing us to recover fine geometry details that were not reconstructed by the Kinect fusion. On the other hand, when a rough mesh is obtained from a single depth map, we apply joint-bilateral filtering~\cite{Kopf07tog} on the depth map to suppress depth noise. As a guidance image for the joint-bilateral filtering, we utilize the corresponding IR shading images.

\subsection{Albedo Estimation}
\label{sec:Albedo}
\noindent{\bf Global Albedo} Since we use IR images, if a target object is made of the same material without different types of colorant, we assume the surface albedo to consist of a single value. This assumption is valid based on our analyses described in Sec. 3.1. Under this assumption, we estimate the surface albedo of vertices globally, using the inversion of \eqnref{eq:IR_projection_model_2}.
Given the measured intensity, $I$, initial normals $\mathbf{n}$ and the initial depth map $d$ from the projected mesh model
and the known lighting direction, $\mathbf{l}$, we can obtain:
\begin{equation}
c\rho = \frac{1}{Z}\sum_{i=1}^{N}\sum_{\substack{m=1,\\ \mathbf{u}_{i,m} \in \mathcal{V}_{i}}}^M\frac{d_{i,m}^2}{\mathbf{n}_{i,m}\cdot \mathbf{l}_{i,m}}I_m(\mathbf{u}_{i,m}),
\label{eq:globalalbedo}
\end{equation}
where $M$ is the total number of shading images, $N$ is the total number of vertices, and $Z$ is a normalization factor. The undesired effect of 
cast shadow and specular saturation is handled by dropping the measurements where intensity values are either too small or too large.

\begin{figure}
    \centering
    \begin{tabular}{@{}c@{ }c@{ }c@{ }c}
        \includegraphics[width=0.96\linewidth]{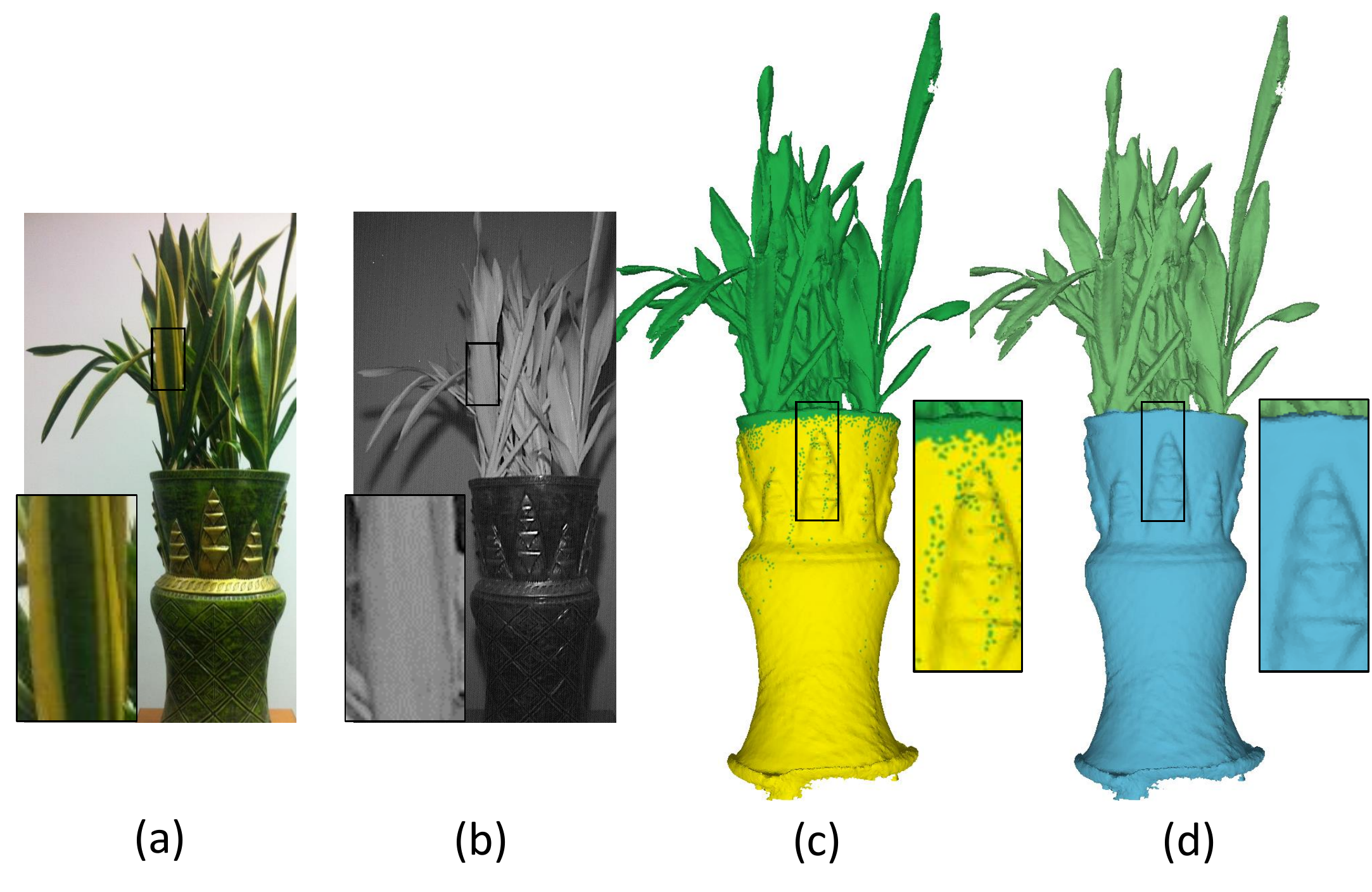}
    \end{tabular}
    \vspace{-3mm}
\caption{Albedo grouping. (a) Color image (b) IR shading image. (c) Color labels of grouped albedo in our previous work ~\cite{Choe14cvpr}. (d) Color labels of grouped albedo with multi-label optimization
}
\label{fig:albedo}
\end{figure}

\vspace{2mm}
\noindent{\bf Multiple Albedo} When a captured object has multiple albedos (multiple materials) in IR images, we compute the albedos on the vertices and group them in the 3D mesh. 
We begin with  estimating the vertex-wise albedos by dividing the captured IR image with the rendered shading image as \eqnref{eq:localalbedo}. 
\begin{equation}
\  c\rho_i = \frac{1}{N_{V_{i}}}\sum_{\substack{m=1,\\ \mathbf{u}_{i,m} \in \mathcal{V}_{i}}}^M\frac{d_{i,m}^2}{\mathbf{n}_{i,m}\cdot \mathbf{l}_{i,m}}I_m(\mathbf{u}_{i,m}),
\label{eq:localalbedo} 
\end{equation}


After estimating the local albedo, we group the local albedos using K-means clustering~\cite{Kanungo02pami} and multi-label optimization~\cite{Boykov04tpami}. Before grouping the albedos, the number of groups, K is decided via principal component analysis (PCA). The dominant directions of feature space are computed and we set K for capturing more than 95\% of the feature space. The feature space consists of vertex positions and local albedos ($\kappa\mathbf{x}_i,c\rho_i$) where the parameter $\kappa$ normalizes the features. After K-means clustering, we improve the albedo grouping via multi-label optimization as follows: 
\begin{equation}
E(p)= \sum_{p=1}^{N}D_p(L_p)+\sum_{p=1}^{N}\sum_{q\in \mathcal{N}_{p}}V_{p,q}(L_p,L_q),
\label{eq:multiopt}
\end{equation}
where $p$ is a vertex index, $q$ are the neighboring vertices of $p$, and $L$ is the label for grouping.
The initial labels from K-means clustering are used for the data term $D$ and we set the neighboring constraint $V$ based on our mesh connectivity.

 \figref{fig:albedo} shows an example of our albedo grouping. In (c), our previous work shows a noisy result which is caused by specularity in the flowerpot. In contrast, we see that the noisy regions are improved and the result becomes more reliable in (d).    
This process gives us a more reliable albedo estimation.



\begin{figure}
    \centering
\vspace{4mm}
    \begin{tabular}{@{}c@{ }c@{ }c@{ }c}
        \includegraphics[width=0.67\linewidth]{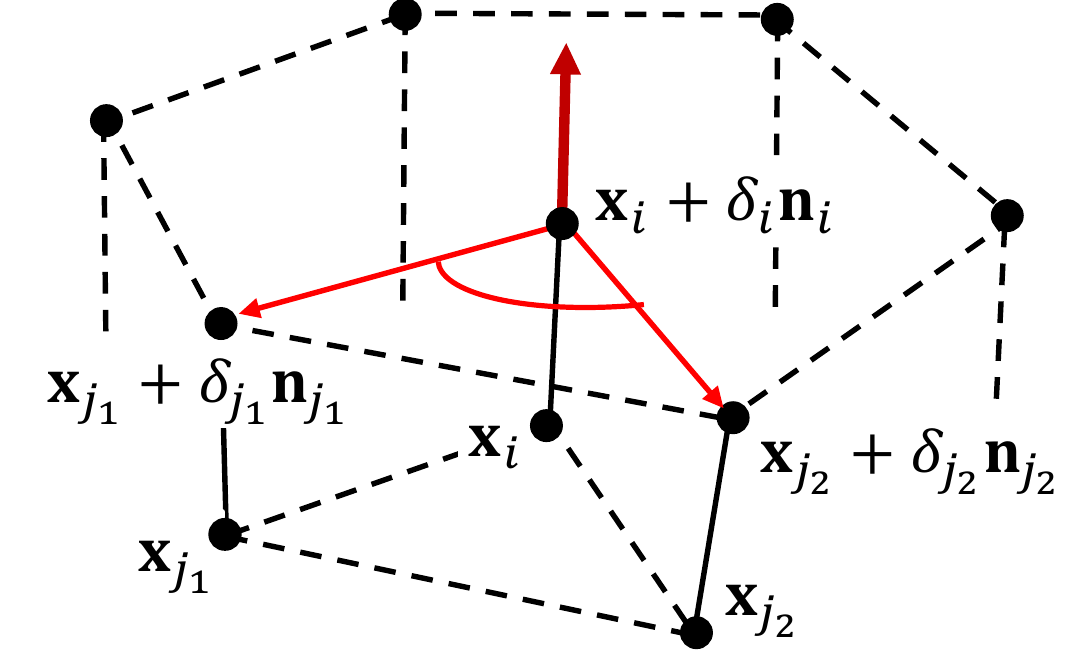}
    \end{tabular}
\caption{Visualization of mesh vertices. Analytic Jacobian of a vertex is defined using the connected neighboring vertices. }
\label{fig:JacobianNormal}
\end{figure}

\subsection{Mesh Optimization}
\label{sec:optimization}
We refine the initial mesh model by searching for the optimal displacement of vertex along its normal direction. The refinement is subject to the shading constraint from the Kinect IR images. We define our cost function as follows:
\begin{eqnarray}
\label{eq:optimization}
&&\arg\min_{\boldsymbol{\delta}} (E_{p}(\boldsymbol{\delta}) + E_{s}(\boldsymbol{\delta})+ E_{r}(\boldsymbol{\delta})), \\
\label{eq:optimizationdata}
E_{p}(\boldsymbol{\delta}) &=& \sum_{i=1}^{N}\sum_{k \in \mathcal{V}_{i}}w_{i,k}\bigg( I_{i,k}-c\rho_{i,k}\frac{\mathbf{n}_{i,k}(\delta_{i,k})\cdot \mathbf{l}_{i,k}}{d_{i,k}^2}\bigg)^{2},\\
\label{eq:optimizationsmoothness}
\!\!\!\!\!\!E_{s}(\boldsymbol{\delta}) &\!\!\!\!\!\!=\!\!\!\!\!\!& \sum_{i=1}^N\sum_{j\in \mathcal{N}_{i}}\lambda_{1}(\delta_i-\delta_j)^2,\\
\label{eq:optimizationregularization}
\!\!\!\!\!\!E_{r}(\boldsymbol{\delta}) &\!\!\!\!\!\!=\!\!\!\!\!\!& \sum_{i=1}^N\lambda_{2}(\delta_i)^2,
\end{eqnarray}
where $\boldsymbol{\delta}=\{\delta_i\}^N_{i=1}$ denotes the displacement of vertices which we want to optimize, and $\bf n_{i,k}$ is the normal direction of the $i$-th vertex projected on the $k$-th view. Our cost function is composed of a data term $E_p(\boldsymbol\delta)$, a smoothness term $E_s(\boldsymbol\delta)$, and a regularization term $E_r(\boldsymbol\delta)$. The relationship among the variables are illustrated in Figure 9.

\begin{figure}
    \centering
    \begin{tabular}{@{}c@{ }c@{ }c@{ }c}
        \includegraphics[width=1\linewidth]{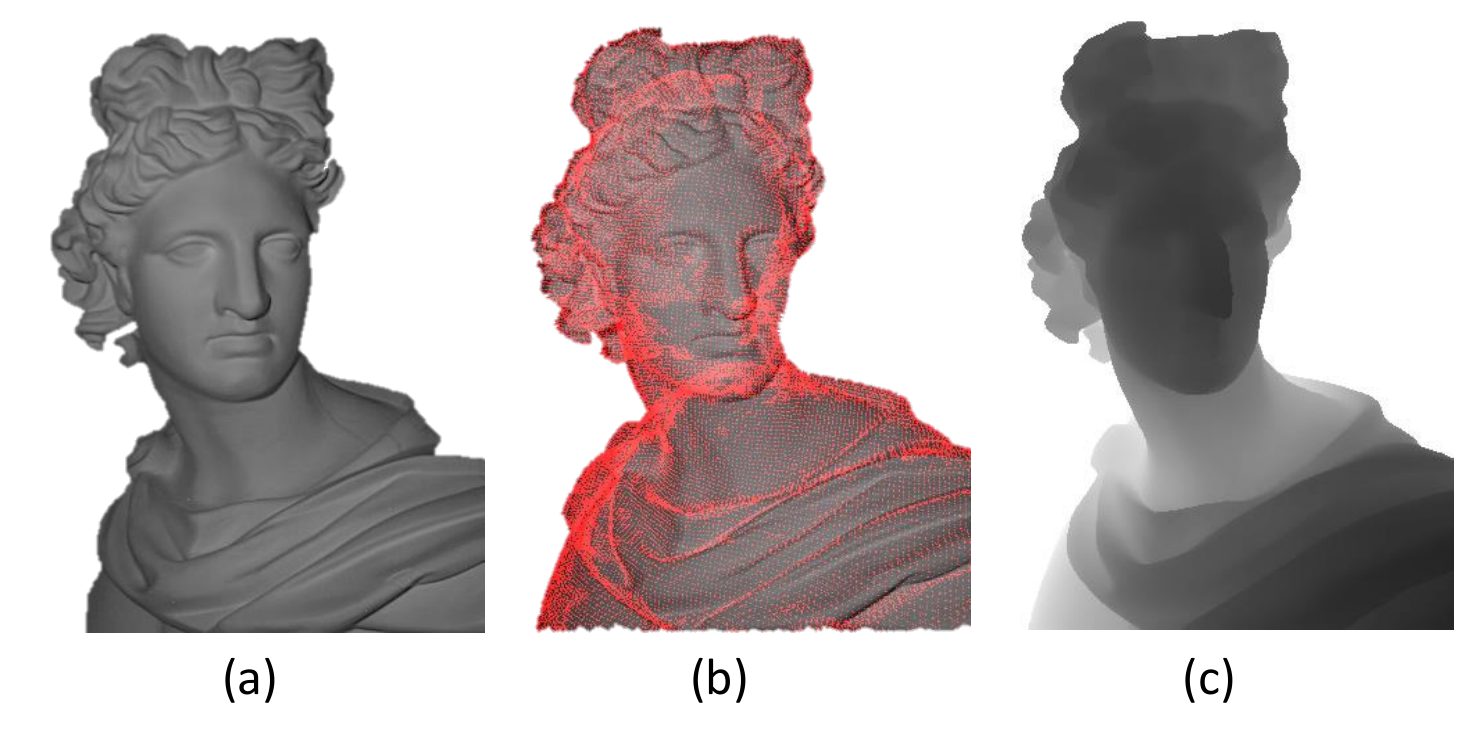}
    \end{tabular}
    \vspace{-2mm}
\caption{(a) One of our input shading images. (b) Projected mesh vertex (red dots) on (a). (c) Depth map derived from a projected mesh model. Note
that the derived depth map from Kinect fusion is far more accurate than the RAW depth map from Kinect. In our geometry refinement process,
we use this depth map instead of the Kinect RAW depth map for mesh optimization.}
\label{fig:vertice_projection}
\end{figure}

The data term $E_{p}(\boldsymbol{\delta})$ in \eqnref{eq:optimizationdata} is designed according to the near light IR shading model described in \secref{sec:LightModel}. At the beginning of our refinement, the IR camera centers are initially estimated in the world coordinate. Since we utilize the calibrated IR camera and the attached light source, the light direction $\mathbf{l}_{i,k}$ at the each light positions can be estimated using the estimated IR camera poses which can be obtained from the Kinect fusion. The distance $d$ between a light source and a vertex position is estimated via the vertex projection, as illustrated in \figref{fig:vertice_projection}. $w_{i,k}$ is the confidence weight expressed by $\mathbf{n}_{i,k}\cdot \mathbf{l}_{i,k}$. Thus, more confidence is given to the vertex which normal direction is closer to the light direction. Since the estimated $d$ is measured in $mm$ and has large effects compared to the other terms, the optimization is sensitive to the depth $d$. Therefore, we begin our optimizing process with the depth-multiplied shading image $I*D$ (The operator* indicates pixel-wise multiplication) where $I_{i}\in I, d_{i} \in D$ and we fix $d$ as a constant at every iteration.

The smoothness term $E_{s}(\boldsymbol{\delta})$ in \eqnref{eq:optimizationsmoothness} modulates the change of displacement which should be locally smooth among the  neighboring vertices. The regularization term $E_{r}(\boldsymbol{\delta})$ in \eqnref{eq:optimizationregularization} regulates the estimated displacement $\delta_{i}$ to be small since the initial mesh from the Kinect fusion is already quite accurate. The $\lambda_1 $ and $\lambda_2$ are manually determined based on the vertex visibility $V$ and mesh scale.

Compared to~\cite{Hernandez08pami}, our method has an advantage to optimize only a single variable $\delta$ for each vertex, which simplifies the optimizing process and makes our process more stable while the method in~\cite{Hernandez08pami} needs to optimize 3 variables,~\ie~x, y, and z displacements for each vertex. By adjusting $\delta_{i}$ of each vertex $\mathbf{x}_{i}$, the position of each vertex $\mathbf{x}_{i}$ is iteratively updated, which minimizes our optimization cost in \eqnref{eq:optimization}. Note that the update of the vertex position for every iteration considers all the shading images at once. We optimize \eqnref{eq:optimization} by utilizing a sparse non-linear least square optimization tool\footnote{SparseLM: Sparse Levenberg-Marquardt nonlinear least squares \url{http://users.ics.forth.gr/~lourakis/sparseLM/}}. At iteration $t$, $\boldsymbol{\delta}$ is determined by minimizing the cost in \eqnref{eq:optimization}, subject to the configuration of vertices at the previous iteration $t-1$.
The iterative update rule for the new vertex location is defined as:
\begin{equation}
\mathbf{x}_{i}^{t} = \mathbf{x}_{i}^{t-1} +  \delta_{i,t} \mathbf{n}_{i}.
\label{eq:vertexupdate}
\end{equation}

After we update the vertices location, the normal directions $\bf n$ are also updated.
In order to solve our objective function efficiently, we derive an analytic Jacobian which provides a deterministic form of $\delta_i$. Given a mesh configuration, in order to estimate $\delta_i$ of a vertex, the objective function in Eq. (5) only requires the location of the connected neighboring vertices to define the smoothness term. The Jacobian matrix of  \eqnref{eq:optimizationdata}, \eqnref{eq:optimizationsmoothness}, and \eqnref{eq:optimizationregularization} are constructed as follows. 




The Jacobian matrix of \eqref{eq:optimizationdata} is:
\begin{equation}
J_{p}(i,j) = \frac{\partial}{\partial\delta_{i}}\bigg( I_{i,k}-c\rho\frac{\mathbf{n}_{i,k}(\delta_{i,k})\cdot \mathbf{l}_{i,k}}{d_{i,k}^2}\bigg)^{2},
\label{eqn:weight_segment}
\end{equation}
where $\mathbf{n}_{i,k}(\delta_{i,k})$ is expressed as:
\begin{equation}
\resizebox{1\hsize}{!}{$\big\{(\mathbf{x}_{i} + \mathbf{\delta}_{i}\mathbf{n}_{i,k})- (\mathbf{x}_{j1} + \mathbf{\delta}_{j1}\mathbf{n}_{j1,k})\big\} \times \big\{({\mathbf{x}_{i} + \mathbf{\delta}_{i}\mathbf{n}_{i,k})- (\mathbf{x}_{j2} + \mathbf{\delta}_{j2}\mathbf{n}_{j2,k})} \big\}$},
\label{eqn:weight_segment}
\end{equation}
the indices 1 and 2 of the neighbor vertices are determined to meet the right-hand rule of the cross product. This guarantees that the direction of $\mathbf{n}_{i,k}(\delta_{i,k})$ is going outward from the mesh, which follows the notation in \figref{fig:JacobianNormal}.\\

Similarly, the Jacobian matrix of \eqref{eq:optimizationsmoothness} is defined as:
\begin{equation}
J_{s}(i,j) =
\left\{
\begin{array}{ll}
-1 & \textrm{if $j \in \mathcal{N}_i$} \\
0 & \textrm{otherwise}, \\
\end{array} \right.
\label{eqn:weight_segment}
\end{equation}


and the Jacobian matrix of Eq.(8) is defined as:

\begin{equation}
J_{r}(i,j) =
\left\{
\begin{array}{ll}
1 & \textrm{if $i = j$} \\
0 & \textrm{otherwise}. \\
\end{array} \right.
\label{eqn:weight_segment}
\end{equation}


The Jacobian matrix $J$ is built by concatenating each of the submatrices  $J_p$, $J_s$ and $J_r$, and the optimal $\boldsymbol{\delta}$ is solved accordingly. As depicted in \secref{sec:ResultComparison}, the analytic Jacobian improves the output quality.
Because our method optimizes vertex positions along with the surface normal direction, if an initial mesh is noisy with uneven surface normal directions, the optimization can easily be trapped in a dissatisfactory solution. With the mesh pre-processing stage in~\secref{sec:preprocessing}, we observe that the optimization produces good results even if we are using the least square form of the cost function.

\bgroup
\def\arraystretch{1.0}
\begin{table}
\centering
\caption{Several commercial depth cameras using near IR band. These belong to one of the two categories : Structured light (SL) and Time-of-Flight (TOF) based.} 
\vspace{-2mm}
\resizebox{1.0\linewidth}{!}{
\begin{tabular}{|c|c|c|c|c|}
\hline
Sensor name & Producer & Type & Resolution & Release \\ \hline
Kinect I & Microsoft & SL & $640\times480$ & 2010 \\ \hline
Xtion Pro Live & Asus & SL & $640\times480$ & 2011 \\ \hline
Carmine & PrimeSense & SL & $640\times480$ & 2013 \\ \hline
RealSense R200 & Intel & SL & $640\times480$ & 2015 \\ \hline 
RealSense F200 & Intel & SL & $640\times480$ & 2015 \\ \hline \hline
Kinect II & Microsoft & TOF & $512\times424$ & 2013 \\ \hline
Senz3D & Creative & TOF & $320\times240$ & 2013  \\ \hline
Pico & PMD & TOF & $160\times120$ & 2013 \\ \hline
DepthSense 536B & SoftKinetic & TOF & $240\times160$ & 2015 \\ \hline
\end{tabular}
}
\label{table:Sensors}
\end{table}
\egroup

\section{Experimental Result}
\label{sec:results}
For the experiments on the Kinect I and II, which are the most representative commercial depth sensor among listed in \tabref{table:Sensors}, we capture 10 to 30 IR shading images with the resolution of $640 \times 480$ and $512 \times 424$, respectively, and used them for our geometry refinement.  We use the Kinect fusion provided in the Kinect SDK 1.7 and 2.0 for estimating initial geometry.  We also validate that our method not only works for multiple image refinement but can also be applied for single image refinement. Result comparisons between the initial and the refined meshes for several challenging real world dataset are provided in \figref{fig:result} and \figref{fig:result2}. The example real world objects we use in this work are: Apollo, Cicero, Towel, Flowerpot, Human face, Ammonite, Sweater, and Ornamental stone model. These examples are made of different types of materials and contain fine geometry details. The fine geometry details were not captured in the RAW Kinect depth maps, nor in the mesh model reconstructed by the Kinect fusion. After applying our geometry refinement, the fine details are recovered in our refined mesh model. We render the mesh models as Phong-shaded models.

\vspace{0mm}
\begin{figure}
\centering
\includegraphics[width=0.95\linewidth]{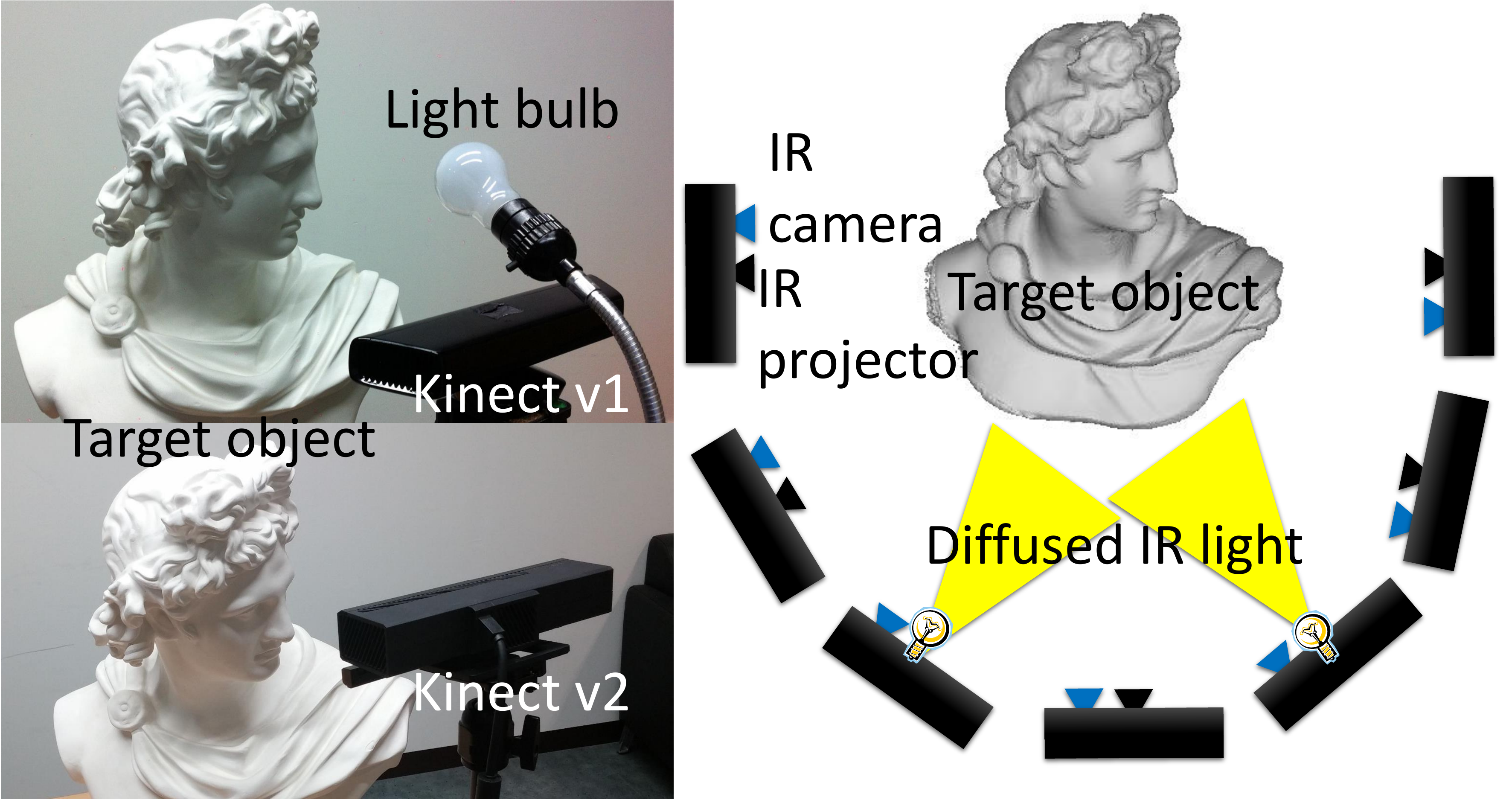}
\caption{Our data capturing system. We use Kinect fusion to obtain an initial base mesh. When utilizing Kinect I, at certain camera positions,
IR camera is blocked and a diffuse light is turned on for capturing shading images. For the Kinect II experiment, since the diffuse IR light source is replaced with the inherent IR projector, additional light bulbs are not used.}
\label{fig:datacapture}
\end{figure}

\subsection{Data Capturing}
Our data capturing process is composed of two main modules, which are the initial geometry acquisition and IR shading image acquisition. \Figref{fig:datacapture} shows our data capturing system. Using Kinect I, we obtain the initial mesh model from Kinect fusion while scanning the target object. At the same time, IR shading images are captured at several discrete viewpoints. When capturing the IR shading images, Kinect fusion is paused to update the mesh, and the Kinect IR projector is blocked so that the uniform IR light constructs our desired IR shading images. We use an additional wide spectrum point light source since we cannot switch the speckle pattern to a uniform IR light from the Kinect IR projector using the Kinect SDK. \footnote{Kinect IR projector is hard-wired and cannot be modifed} Note that this process can be simplified by using a Kinect IR projector if the pattern from the Kinect IR projector is programmable. The locations where we capture shading images belong to the subset of camera poses during Kinect fusion. The camera poses are estimated using the Kinect SDK by registering Kinect depth map with the current reconstructed surface. The relative location of the additional wide spectrum point light source and the Kinect IR camera is fixed and pre-calibrated. Therefore, lighting direction, $\mathbf{l}$ in \eqnref{eq:optimizationdata}, is known after data capturing. 

The capturing process of Kinect II takes the same form as that of the Kinect I. However, the Kinect II emits a uniform IR light and does not require the additional light source, which makes our setup simpler. Additionally, we capture a depth and IR shading image pair at the single viewpoint for further analysis. Since the indoor ambient lights does not affect the captured IR image, both data acquisition is performed under natural indoor lighting.

\subsection{Qualitative Evaluation}
\label{sec:ResultComparison}
We compare the geometries obtained from Kinect fusion and our refined results on the real-world objects that exhibit different shading and albedo characteristics.  Also, we analyze the effect of using analytic Jacobian and the difference of using multiple and single image.

\begin{figure*}
\vspace{-0.15in}
    \centering
    \begin{tabular}{@{}c@{ }c@{ }c@{ }c}
        \includegraphics[width=1\linewidth]{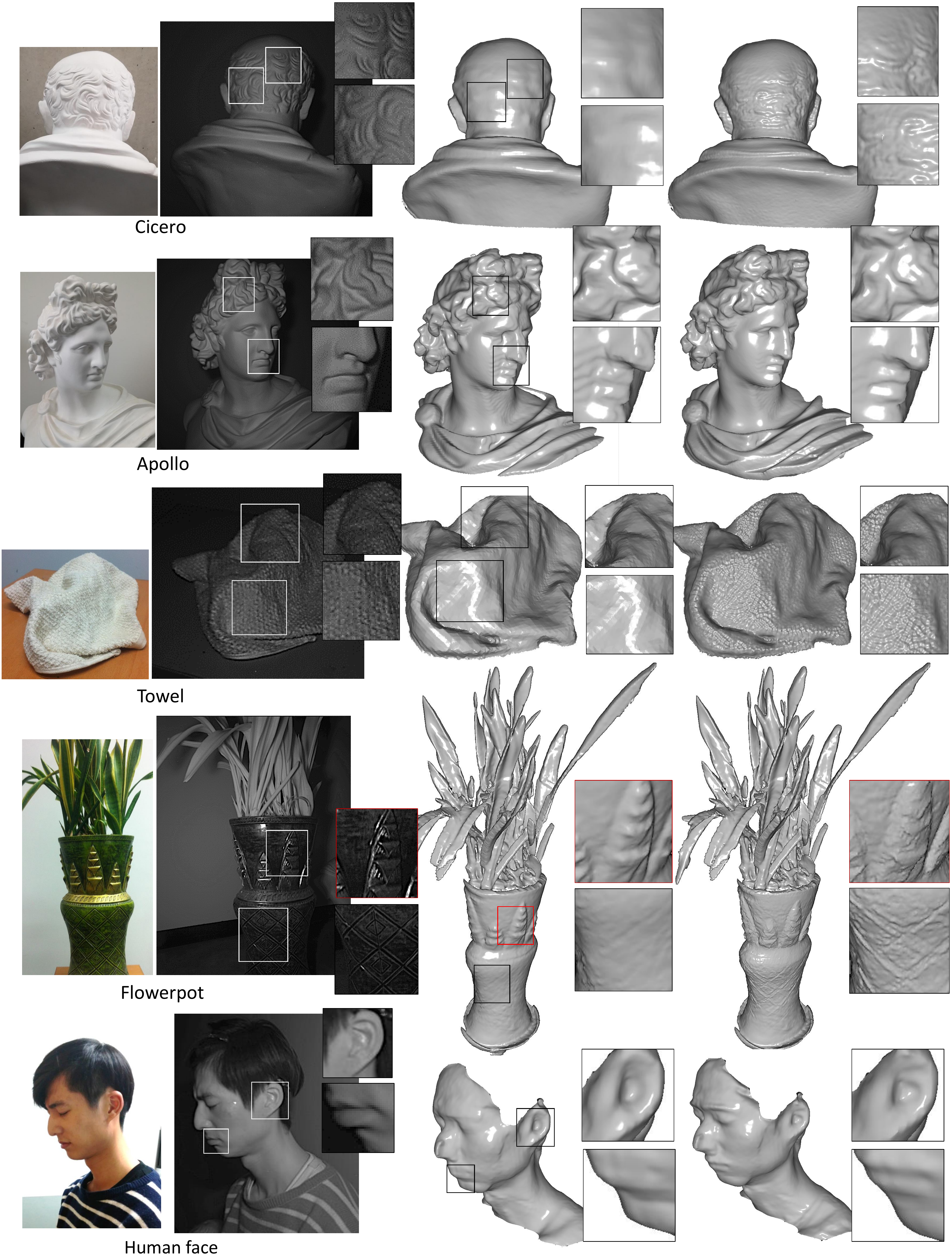}
    \end{tabular}
    \caption{Result comparison of real world objects - Apollo, Cicero, Towel, Flowerpot and Human face. From the left, each column represents color images, IR shading images, initial mesh from Kinect fusion and our mesh result, respectively. Note that our method only requires IR shading images for geometry refinement and the color images are shown for the visual comparison with our IR shading images. }
\label{fig:result}
\end{figure*}

\begin{figure*}
    \centering
    \begin{tabular}{@{}c@{ }c@{ }c@{ }c}
        \includegraphics[width=0.99\linewidth]{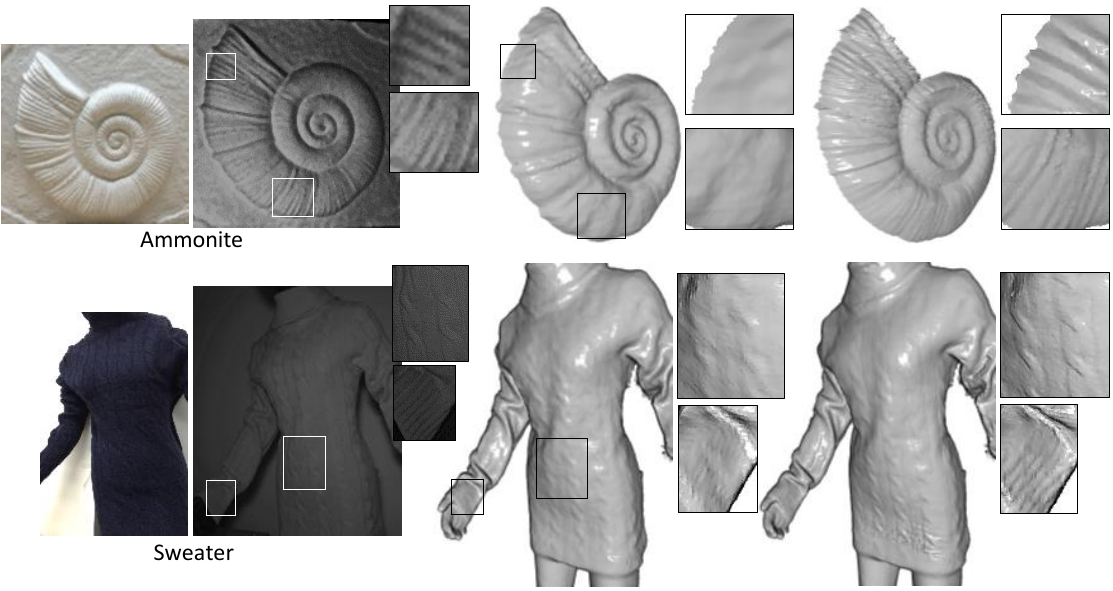}
    \end{tabular}
\caption{Result comparison of real world objects - Ammonite (Obtained from Kinect II) and Sweater. From the left, each column represents color images, IR shading images, initial mesh from Kinect fusion and our mesh result, respectively. Our method only requires IR shading images for geometry refinement and the color images are shown for the visual comparison with our IR shading images. }
\label{fig:result2}\vspace{-0.1in}
\end{figure*}

\vspace{2mm}
\noindent{\bf Cicero}
{
The statue of Cicero is made of plaster and has fine geometric details on its face and hair region. The size of Cicero is $0.7m\times 0.45m$. In \figref{fig:result}, the initial mesh from Kinect fusion and enhanced mesh from our method are compared.
The back of Cicero's head exhibits very fine levels of detail that are not shown in the initial mesh at all. In our result, the fine hair details are recovered. 22 IR shading images are used here. We provided an additional comparison with RGB shading-based refinement method proposed by Han \etal, \cite{Han13ICCV} in \figref{fig:CompareJY}. The color based approaches need to encode the surrounding light environment if the image is not taken using the point light source in a dark room condition. These approaches involve spherical harmonic or polynomial environment light representation. Whereas, the benefit of IR image is that it is like a darkroom photo and initial geometry can be refined even if simple near light source model is applied. As shown in \figref{fig:CompareJY}, the refined mesh using our approach is comparable to the color based approach.
We provide the 3D models that scans complete 360 degree view of Cicero in \url{http://rcv.kaist.ac.kr/gmchoe/project/Kinect_IR/}

}

\vspace{2mm}
\noindent{\bf Apollo}
{
A statue of Apollo (size of $0.75m\times 0.65m$) is also used to verify our algorithm.
The IR shading image shows that Apollo has a double eyelid on its eye but it is not expressed in the mesh from Kinect fusion. Apollo also has fine details for its hairs but were not conveyed in the initial mesh. Our refinement on the initial mesh shows enhanced double eyelids and hair geometry. We used 24 IR shading images for the result. 
}

\vspace{2mm}
\noindent{\bf Towel}
{
We verified that our method works well on small objects with subtle details. A towel, size of $0.2m\times 0.2m$ , was used for our experiment. As shown in \figref{fig:result}, result of towel, initial mesh loses its fine, checkered pattern and shows a flat surface geometry. However, our method can effectively recover the checkered pattern in detail and the surface of our result mesh becomes rather similar to the geometry of the real object.
}

\vspace{2mm}
\noindent{\bf Flowerpot}
{
We tested our algorithm with a multi-albedo object. The target object is a plant with a pot, measuring at $1.2m\times 0.3m$. We grouped the albedo as described in \secref{sec:Albedo}. As shown in \figref{fig:albedo}, plant leaves and the pot have different observation in surface albedo in the IR image. We observed that the plant leaves have smooth geometry and there was less room for refining geometric details. On the other hand, the pot has a complex geometry. We apply our method on the initial mesh from Kinect fusion. In this case, our method for multi-albedo object in Sec.~\ref{sec:Albedo} is applied prior to the mesh optimization. The cross stripes on the pot are recovered by using our method. However, the region that is marked with the red box shows less reliable result. In this region, specularity exists and it does not follow the Lambertian shading model in \eqnref{eq:IR_projection_model_2}.
}

\vspace{2mm}
\noindent{\bf Human face}
{
Our method shows better mesh results for human faces as well. we captured the initial geometry and IR shading images moving around the face while the subject fixed his position and facial expression. 
For this experiment, we use 7 IR images to refine the 3D model. 
We see that the refined result shows more details at the eyes, lips, and ears compared to the mesh from Kinect fusion. Two facial models are used and evaluated. 
}

\begin{figure*}
    \centering
    \begin{tabular}{@{}c@{ }c@{ }c@{ }c}
        \includegraphics[width=0.99\linewidth]{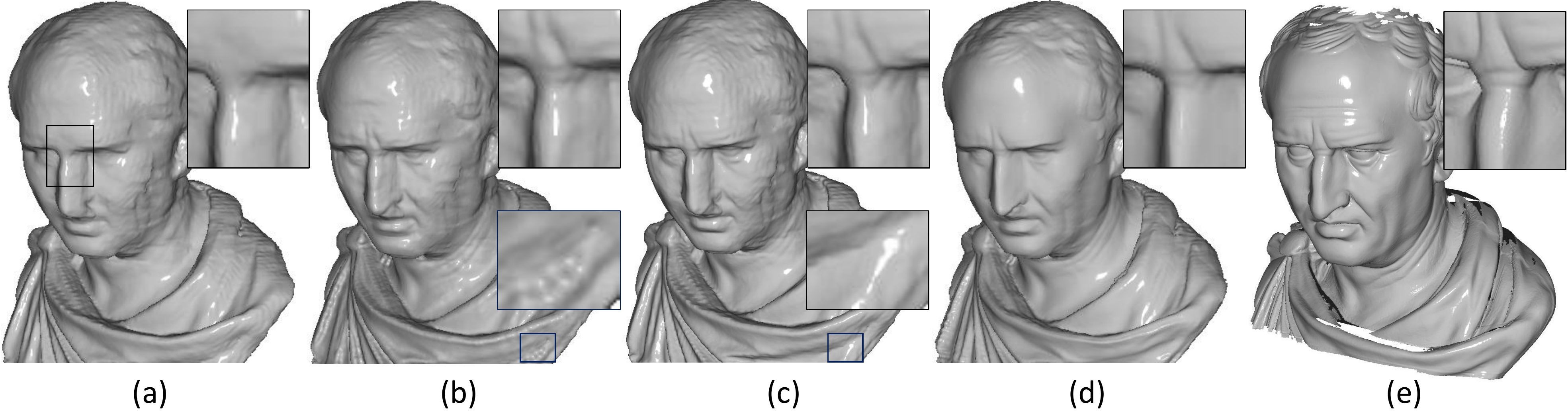}
    \end{tabular}
\caption{Result comparison of real-world dataset. (a) Initial mesh model (b) Our result using a single shading image via numerical Jacobian optimization. (c) Our result using a single shading image via analytic Jacobian optimization. (d) Our result using 36 shading images. (e) Ground truth generated from a structured-light based 3D scanner. In (b), wave-like artifact is shown. On the other hand, the wave-like artifact is suppressed in (c), which shows better convergence of the optimization using the analytic Jacobian. Average distance error of (a) and (d) w.r.t the ground truth model (e) are 2.041 and 2.010mm respectively. }
\label{fig:CiceroResultCompare}
\end{figure*}

\begin{figure*}
    \centering
    \begin{tabular}{@{}c@{ }c@{ }c@{ }c}
        \includegraphics[width=0.99\linewidth]{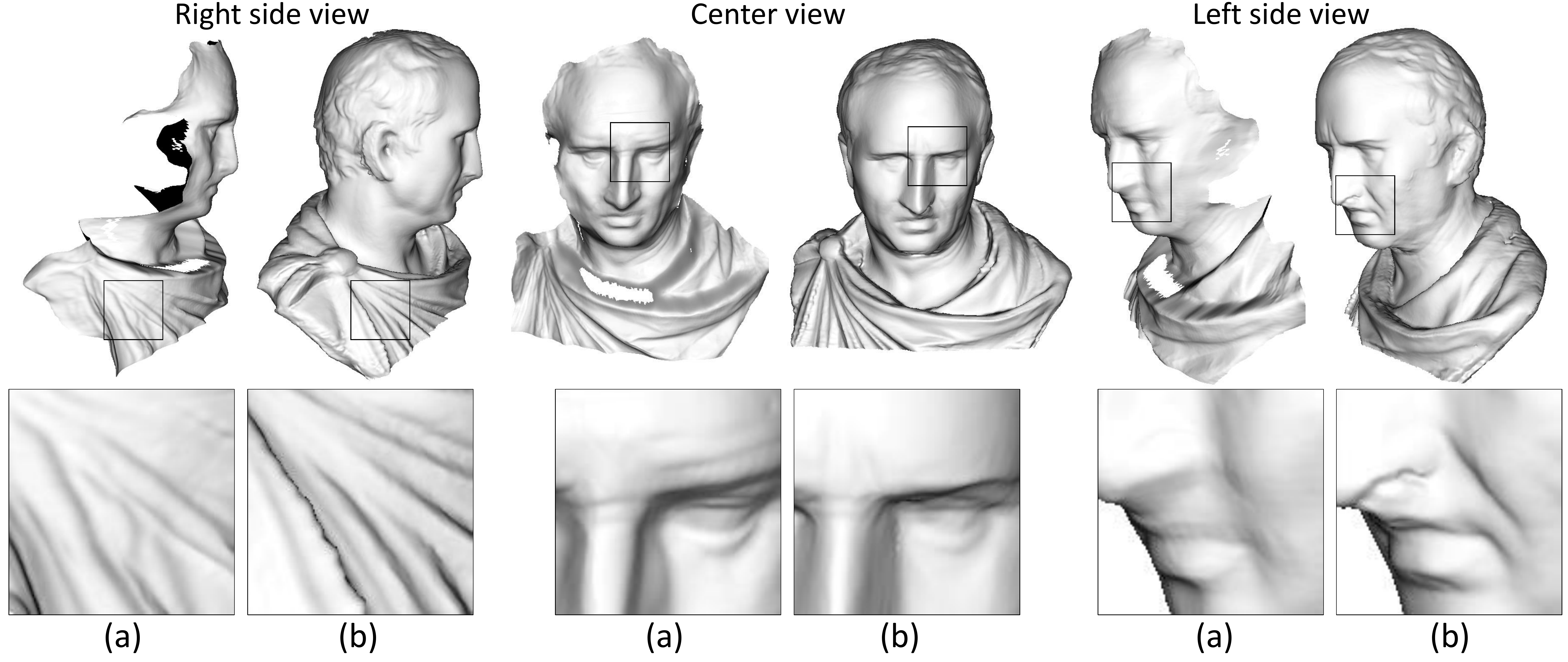}
    \end{tabular}
\caption{Comparisons of results with the conventional method, \cite{Han13ICCV}. Three different view points are compared. (a) Refined mesh result from \cite{Han13ICCV}.  (b) Our mesh result. Our method works better for all-around views. Even if simple near light source model is applied, our approach is comparable to the color based approach. }
\label{fig:CompareJY}\vspace{-0.1in}
\end{figure*}

\vspace{2mm}
\noindent{\bf Ammonite}
{
Ammonite is made of plaster and is a relief sculpture with one side of the plane is carved similar to an ammonite fossil.
The size of the foreground object is $0.24m\times 0.23m$.  
The structure of an ammonite shell is planispiral with very fine stripe patterns.
Since a depth difference between the adjacent patterns is less than 1mm, we see it can not be captured from Kinect fusion mesh. 
However the captured IR shading image shows the original shape containing the fine stripe patterns on it and our result is optimized to exactly follow the real geometry. 
To refine this mesh, 3 IR shading images are used. 
}

\begin{figure*}
\vspace{-0.2mm}
    \centering
    \begin{tabular}{@{}c@{ }c@{ }c@{ }c}
        \includegraphics[width=1\linewidth]{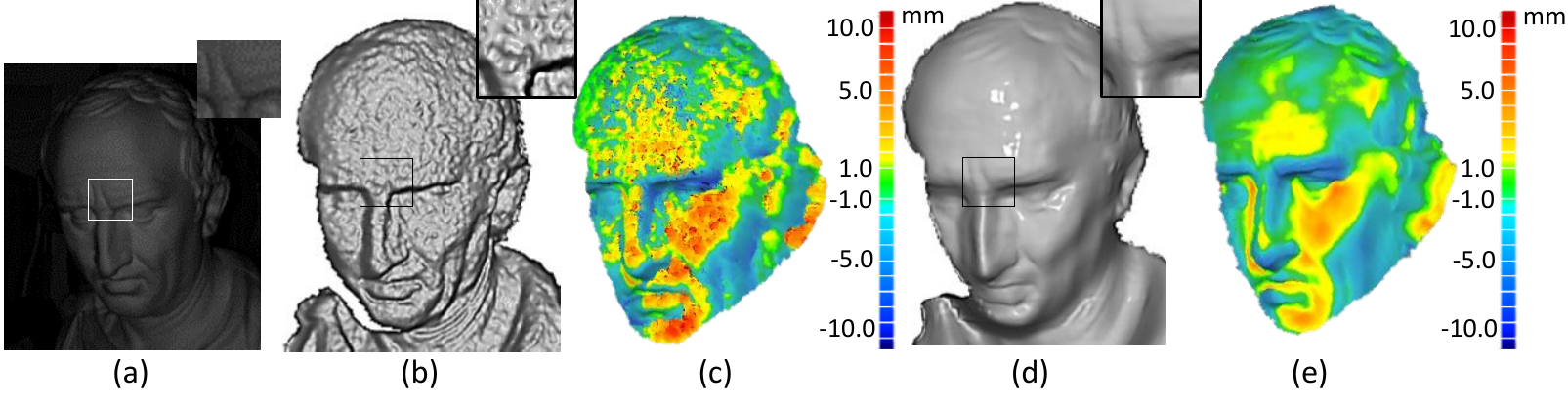}
    \end{tabular}
    \vspace{-0.2in}
    \caption{Result comparison of Cicero dataset captured from Kinect II. By aligning the meshes to the ground truth model obtained from structured-light scanner, we compute metric error of the initial geometry from Kinect and our refined geometry. (a) IR shading image. (b) Kinect II raw depth. (c) Visualization of the metric error of (b). (d) Our refined result. (e) Visualization of the metric error of (d)}
\label{fig:ResultCiceroKinect2}
\end{figure*}

\vspace{2mm}
\noindent{\bf Sweater}
{
Sweater is made of wool and has repetitive twisted patterns on it. It is $0.8m$ high and $0.4m$ wide. The measured depth variation of the twisted pattern is $1mm$. The second row of \figref{fig:result2} shows the IR shading image, initial mesh, and our results for the sweater dataset. The geometry from Kinect fusion does not fully express the twisted pattern on the sweater. On the other hand, our result recovers the twisted pattern clearly.
}

\begin{figure*}[t]
\vspace{1mm}
    \centering
    \begin{tabular}{@{}c@{ }c@{ }c@{ }c}
        \includegraphics[width=1\linewidth]{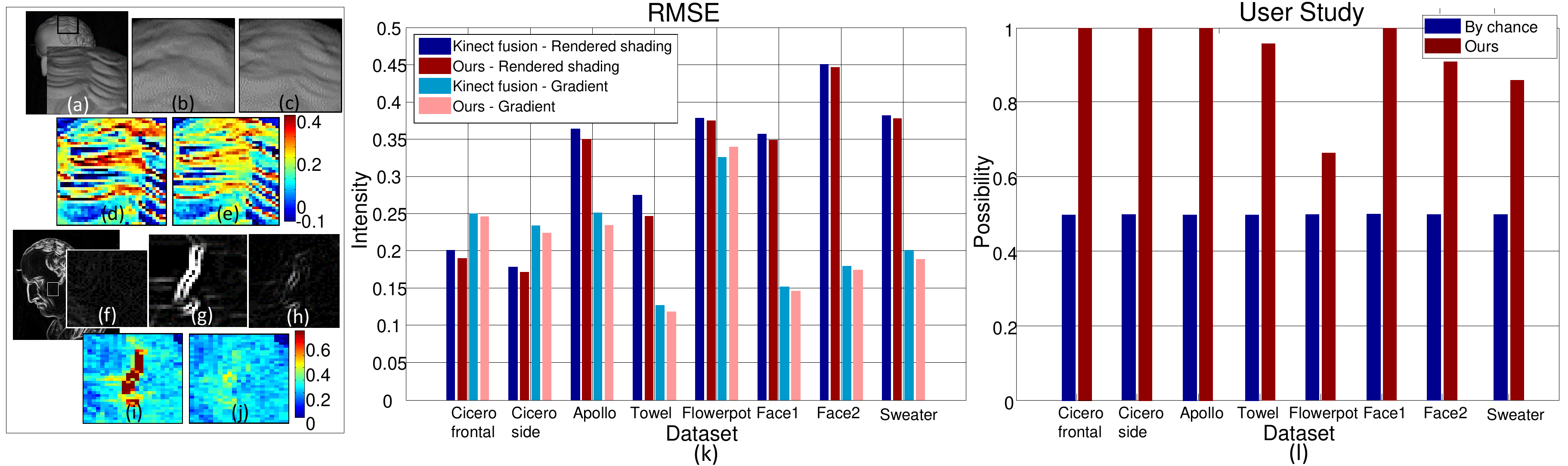}
    \end{tabular}
    \caption{Rendering errors and the user study results. (a) IR shading image. (b, c) Rendered images from Kinect fusion and ours. (d) Error map of (a) and (b). (e) Error map of (a) and (c). (f, g, h) Gradient images of (a, b, c) respectively. (i) Error map of (f) and (g). (j) Error map of (f) and (h). (k) RSME chart. (l) User-study chart .}
\label{fig:ResultQuanti}
\end{figure*}

\vspace{2mm}
\noindent{\bf Effect of Analytic Jacobian }
{
As our approach applies optimization for mesh refinement, the analytic Jacobian described in~\secref{sec:optimization} is helpful for an efficient optimization. To verify the effect, we utilize both numerical Jacobian and analytic Jacobian for the mesh optimization using Cicero dataset. The result is shown in \figref{fig:CiceroResultCompare}. For each of the experiments, $\lambda_{1}$ and $\lambda_{2}$ are set to be optimal. In \figref{fig:CiceroResultCompare}~(b, c), wrinkles of the forehead and eyes are refined well in both cases (see upper bound box). However, in the neck and the torso region of the model, the two cases show differences in terms of its quality. In \figref{fig:CiceroResultCompare}~(b), some wave-like artifact is caused. On the other hand, \figref{fig:CiceroResultCompare}~(d) shows better results for the refined mesh, as the wave artifact is suppressed (see lower bound box in the figure). For each cases, mean errors of our cost function is computed after the refinement. The case of using analytic Jacobian shows less error.      

}

\vspace{1mm}
\noindent{\bf Number of  Images }
{
As depicted in \eqnref{eq:optimization}, IR shading images are used for giving photometric cues to each vertex. According to the number of the input IR shading images, the quality of the refined mesh shows a difference. \Figref{fig:CiceroResultCompare} compares (a) inital mesh from Kinect fusion, (c) refined mesh using a single IR shading image and (d) refined mesh using multiple (36) images. Mesh in (c) and (d) show enhanced results where detailed features such as wrinkles in the middle of the forehead and hair are reconstructed. Also, compared to the initial mesh (a), which shows an unsharp nose caused by the loop-closing error of Kinect fusion, our method greatly suppresses errors and reconstruct the original sharpness of geometry in the real world. In (c), however, there still remains some bumpy surfaces on the face and cheek area, same as the initial (a). On the other hand, multiple images refine the surface clearly in (d). Since our method tries to optimize each vertices toward satisfying the IR shading observation, usage of multiple images better solves the shading-geometry ambiguity. We can see a more smooth surface when using multiple shading images.
}

\vspace{-2mm}
\subsection{Quantitative Evaluation}
{To verify that the rendered intensities from refined geometry follows IR shading image better, error measures of initial and our mesh in the image domain is  conducted. Both initial and refined mesh are rendered in image domain based on the \eqnref{eq:IR_projection_model_2}. We also generate first order gradient of rendered image to evaluate how the geometric edges follows the edge in the IR image. We use root mean square error (RMSE) which is equivalent to error between input image and rendered images.
RMSE:= $\sqrt{       \frac       {\sum^n_{t=1}{(I_{in,t}-I_{r,t})}^2}{n}           }$ , where $n$ is pixel number, $I_{in}$ is input shading image and $I_r$ is the rendered image.
To make the evaluation not biased to specific image, we conduct experiment as follows. 1) Among a set of input images, one random image is intentionally omitted. 2) Perform mesh refinement using the non-omitted images. 3) Render an image with novel viewpoint that are equivalent to the viewpoint of the omitted image. 4) Compute RMSE between rendered image and omitted image. In this way, we plotted the bar chart in \figref{fig:ResultQuanti}. According to the bar chart, the error is decreased.

We also compute metric error of the initial geometry and our refined geometry. The ground truth model is obtained from a structured-light based 3D scanner. Using the Iterative Closest Point (ICP) algorithm in~\cite{Besl92tpami}, the meshes are registered to ground truth. Then we compute metric error, which is visualized in \figref{fig:CiceroResultCompare} and \figref{fig:ResultCiceroKinect2}.   

\vspace{2mm}
\noindent{\bf User Study}
Work in \cite{Shan133dv} proposes the visual turing test via user study to evaluate the visual quality of their result. To evaluate the realism of our enhanced 3D mesh model, we conducted a series of user studies. We collected 21 subjects who are not experts of 3D computer vision. For every real-world dataset which we deal with in this paper, the subjects are asked which mesh model between the Kinect fusion and ours is more similar-looking to input IR shading image.
The red bar charts in \figref{fig:ResultQuanti}, (l) show the possibility that our mesh to be responded as a better quality than that of Kinect fusion. The by-chance possibility is $0.5$ for every dataset, which is expressed with blue bars. We see most of the people responded our results are better.

\subsection{Failure Case}
Although we show that our method can refine single depth-IR image of the Cicero dataset, we found that the single image input does not fully guarantee the success of refinement due to shading-geometry ambiguity. In \figref{fig:ResultOrnamentalStone}, a result comparison between an initial geometry and refined geometry for an ornamental stone dataset is shown. The ornamental stone dataset has fine details and it is not represented in the initial geometry. A result from our method (See \figref{fig:ResultOrnamentalStone}~(c)) shows better quality of geometry, whose geometric details follow the input IR shading image. However, when we look at the geometry at different viewpoints, the geometry shows a bumpy surface and less accurate result.  We let this problem as a future work.

\begin{figure*}
\centering
\includegraphics[width=0.99\linewidth]{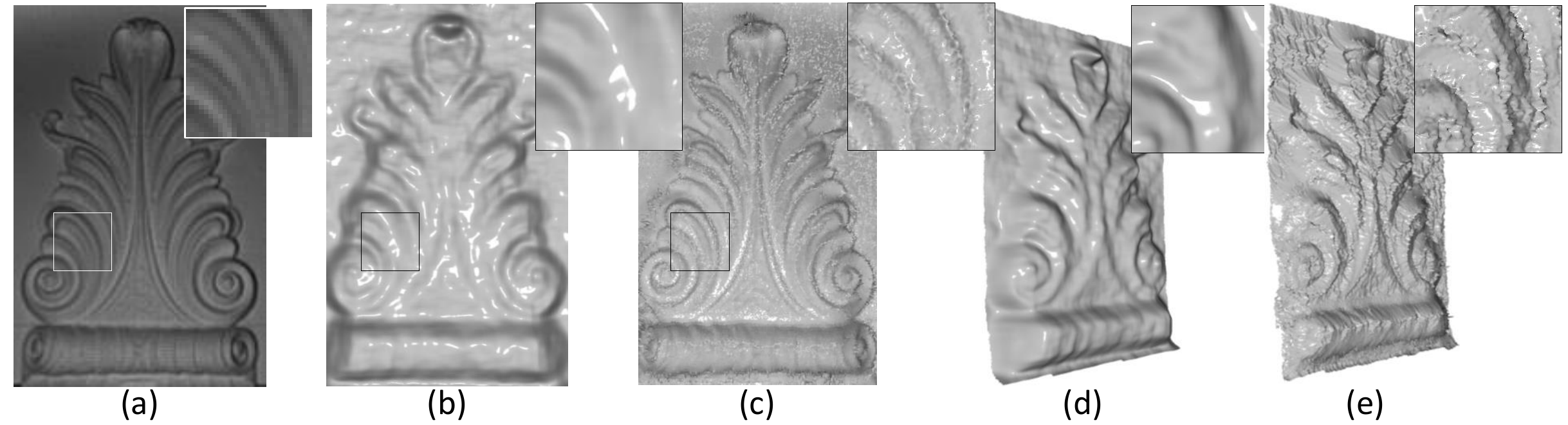}    
\caption{The refinement using single shading image can sometimes be biased. We show the refined result of Ornamental stone dataset obtained from Kinect II. (a) Input IR shading image. (b),(d) Mesh from single depth. (c, e) Our result. Although our method guide the initial mesh to follow the IR shading image, at some different viewpoint, we see unsatisfactory result. }
\label{fig:ResultOrnamentalStone}
\end{figure*}

\section{Discussion}
As a limitation of our work, we assume the Lambertian BRDF which makes our results error-prone to specular highlight. Due to the usage of Kinect fusion algorithm, we also assume the reconstructed object is static. In future, we will study how to extend our work to handle non-Lambertian BRDF objects, and geometry refinement for dynamic object reconstructions. 
The depth based camera tracking is not perfect due to accumulation error of estimated camera poses. Such problem results in unpleasant geometric seams as shown in \figref{fig:CiceroResultCompare} (a). Our algorithm does not target bundle adjustment of camera poses. However, if the amount of tracking error is not severe, our approach can refine geometry to minimize multi-view shading inconsistencies. As shown in \figref{fig:CiceroResultCompare} (d), the refined mesh shows relieved geometric seams and geometric details. We believe this result supports that our approach correctly minimizes the gap between initial geometry and observed shading image even in presence of camera tracking error. 
	For the every results displayed in the paper, we did not process camera poses before mesh refinement. However, if the tracking error is not ignorable, the projection matrices or image coordinate can be further optimized so that the depth and shading images more precisely be aligned as introduced in \cite{zhou2014color,zollhofershading}. About the radiometric calibration, in Chatterjee \etal\cite{chatterjee2015photometric}, they utilize two auxiliary light sources and finds out linearity of the response function. However, according to our repeated experiment, the gamma curve does not fitted to 1 which indicates linear response. We could not exactly reproduce the approach as the paper does not describe which Kinect device is used and how the IR images are grabbed (we utilized Microsoft Kinect SDK 1.7 for Kinect I and 2.0 for Kinect II). However, we agree that shape of response function is near to linear as we seen in\figref{fig:calibration} (b),(c). Here, we choose gamma function as a camera response function because the gamma curve expresses most of the observed intensities fairly well. However, this also opens interesting research direction since the radiometric calibration on the IR cameras is rarely studied compared to the color cameras. About the multiple albedo, our method is built upon simple image formation model assuming constant albedo and Lambertian shading on the scene. Although our extension to care multiple albedo have been demonstrated on the several real-world examples, there is a room for improving our approach to handle complex cases such as non-Lambertian objects exhibiting sub-space scattering, non-constant albedo, or strong specular. Moreover, an effective specular handling mehod should be further studied for enhancing the mesh quality of reflexible objects.  Also, as we analyzed in \figref{fig:ResultOrnamentalStone}, we will try to reinforce our method to more robustly handle the single image refinement.

\vspace{0mm}
\section{Conclusion}
In this paper, we have presented a framework to utilize shading information from Kinect IR images for geometry refinement. This work studies the shading information inherent in the Kinect IR images and utilizes 
them for geometry refinement. As demonstrated in our study, the captured spectrum of Kinect IR images does not have any overlapping with 
visible spectrum which makes our acquisition unaffected by indoor illumination condition. Since there is almost no ambient light in 
IR spectrum, the captured intensity can be accurately modeled by our near light IR shading model assuming the captured materials follow the
Lambertian BRDF. 

We have also described a method to radiometrically calibrate the Kinect IR image using a diffuse sphere, a method to estimate
albedo and do albedo grouping, and a new mesh optimization method to refine geometry by estimating a displacement vector along 
vertex normal direction. Our experimental results show that our framework is effective and demonstrates high-quality mesh model via 
our geometry refinements. Major experiments are done using multiple IR shading images at different viewpoints. The effectiveness of our method is demonstrated via various real-world examples using both Kinect I and Kinect II.


\bibliographystyle{spmpsci}      
\bibliography{egbib}   

%
%

\end{document}